\documentclass[a4paper,fleqn]{cas-dc}

\usepackage[authoryear,longnamesfirst]{natbib}
\usepackage[figuresright]{rotating}
\hypersetup{
  colorlinks,
  citecolor=Blue,
  linkcolor=Blue,
  urlcolor=Blue}
\usepackage{algorithmic}
\usepackage{booktabs}
\usepackage{graphicx}
\usepackage{textcomp}
\usepackage{xcolor}
\usepackage{color}
\usepackage{multirow}
\usepackage{caption}
\usepackage{graphicx}
\usepackage{subfigure}

\def\tsc#1{\csdef{#1}{\textsc{\lowercase{#1}}\xspace}}
\tsc{WGM}
\tsc{QE}

\begin{document}
\let\WriteBookmarks\relax
\def\floatpagepagefraction{1}
\def\textpagefraction{.001}

\captionsetup[figure]{labelfont={bf},labelformat={default},labelsep=period,name={Fig.}}

\shorttitle{Region-Aware Network}    

\shortauthors{Yuehai Chen \emph{et al.}}  

\title [mode = title]{Region-Aware Network: Model Human's Top-Down Visual Perception Mechanism for Crowd Counting}  

\tnotemark[1]


%

\author[1]{Yuehai Chen}[orcid=0000-0002-4778-8718]


\fnmark[1]

\ead{cyh0518@stu.xjtu.edu.cn}



\affiliation[1]{organization={School of Automation Science and Engineering,Faculty of Electronic and Information Engineering},
            addressline={Xi'an Jiaotong University}, 
            city={Xi'an},
            postcode={710049}, 
            state={Shanxi},
            country={China}}

\author[1,2]{Jing Yang}
\fnmark[1]
\ead{jasmine1976@xjtu.edu.cn}


\fntext[cor1]{Yuehai Chen and Jing Yang contributed equally to this work.}



\author[1]{Dong Zhang}
\author[1]{Kun Zhang}

\author[2]{Badong Chen}
\author[2]{Shaoyi Du}
\cormark[1]
\ead{dushaoyi@xjtu.edu.cn}
\affiliation[2]{organization={Institute of Articial Intelligence and Robotics, College of Articial Intelligence},
            addressline={Xi'an Jiaotong University}, 
            city={Xi'an},
            postcode={710049}, 
            state={Shanxi},
            country={China}}


\cortext[1]{Corresponding author at: Institute of Articial Intelligence and Robotics, College of Articial Intelligence.}



\begin{abstract}
Background noise and scale variation are common problems that have been long recognized in crowd counting. Humans glance at a crowd image and instantly know the approximate number of human and where they are through attention the crowd regions and the congestion degree of crowd regions with a global receptive filed. Hence, in this paper, we propose a novel feedback network with Region-Aware block called RANet by modeling human's Top-Down visual perception mechanism. Firstly, we introduce a feedback architecture to generate priority maps that provide prior about candidate crowd regions in input images. The prior enables the RANet pay more attention to crowd regions. Then we design Region-Aware block that could adaptively encode the contextual information into input images through global receptive field. More specifically, we scan the whole input images and its priority maps in the form of column vector to obtain a relevance matrix estimating their similarity. The relevance matrix obtained would be utilized to build global relationships between pixels. Our method outperforms state-of-the-art crowd counting methods on several public datasets.
\end{abstract}




\maketitle

\section{Introduction}
\label{1}

Crowd counting is an essential work in the field of computer vision. It has wide range of applications such as video surveillance, urban planning, public safety \emph{\emph{et al.}} For example, with the rapid growth of population and urbanization, the situations of crowd gathering such as stadium, concerts and parades are more and more frequent. In these scenarios, crowd counting plays an indispensable role for public safety \citep{b60}.  

Although crowd counting task is important and useful, the real usage remains limited since the dense crowd counting is challenging. One of the main challenges is background noise. The influence of complex and irrelevant background is not conducive to correctly identify the crowd. Scale variation may also hurt the counting performance. Since the scale of people varies dramatically in images and across different images, it is hard to extract effective features for density regression. 

To alleviate noises caused by cluttered backgrounds, attention mechanism is usually introduced to focus on crowd regions \citep{b2,b3,b5,b27,b39}. For instance, \citep{b2} proposed an attention-injective deformable convolutional network to generate crowd regions and congestion priors . With similar idea, \citep{b3} presented a novel attention network by incorporating attention maps to better focus on the crowd area .
 
Although these attention-based methods have shown great success in dealing with issue about background noise, they ignore the context information accounting for scale variation in congested scenes by using pixel-wise product of image and corresponding attention map.

Meanwhile, as the heart of these CNNs-based approaches, standard convolutions always exploit the same filters and pooling operations over the whole image to obtain multi-scale features. This operation leads to the situation that the actual size of receptive fields in the networks is much smaller than the theoretical size \citep{b1}. It means that normal CNN networks may have limited effect for rapid scale changes in complex scenes as they may assign a fixed scale for large objects \citep{b22}. 

\begin{figure}
\centering
\centerline{\includegraphics[scale=0.28]{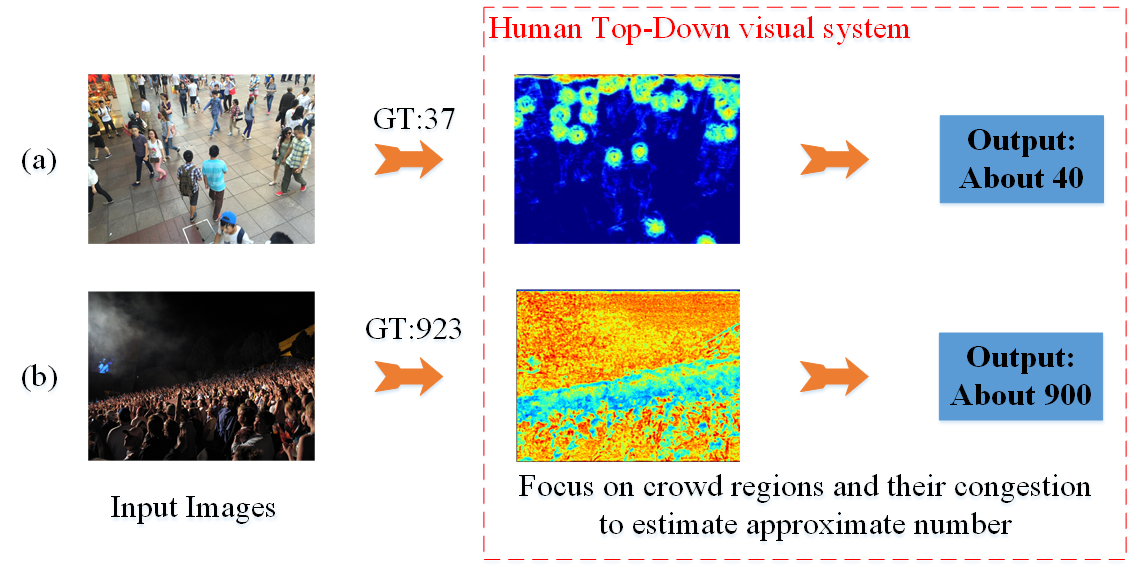}}
\caption{Human's Top-Down visual system: According to the goal of the current task and previous prior knowledge, human scan a crowd image and instantly know the approximate number of human and where they are through attention the crowd regions and the congestion degree of crowd regions with a global receptive filed. The color is brighter, the crowd congestion degree is higher. 'GT' means Ground Truth.}
\label{fig1}
\end{figure}

People's Top-Down visual perception mechanism can solve these above problems well. As Fig. \ref{fig1} shows, when human observer crowd scenes, they quickly scan the whole crowd scenes and focus on the crowd regions based on the prior knowledge. It is conducive to reduce the interference of background noise. Then human estimate the approximate number in crowd scenes through the congestion degree of crowd regions and global receptive fields accounting for scale variation.

Inspired by the human’s Top-Down visual perception mechanism, we propose a novel method to address the issues above for precise crowd counting in this paper. First, we introduce a feedback architecture to generate priority map focusing on crowd regions for reducing interference caused by the background noise. Then, we design a Region-Aware block to adaptively encoder context information for understanding scale variation through global receptive field. In the Region-Aware block, we first do a similarity measurement between the input images and corresponding priority maps through scanning them in the form of column vector. Then, the obtained similarity measurement matrix will be embedded into input images to enhance the crowd regions and build global context relationship accounting for regional consistencies and scale variation.

The main contributions of this paper are as follows:
\begin{itemize}
\item We exploit the feature information from priority maps to focus on crowd regions for reducing interference caused by the background noise.
\item Through the designed Region-Aware block, the network can encoder the context information and expand the size of receptive field for understanding scale variation.
\item With the proposed framework, we achieve state-of-the-art performance on most crowd counting benchmarks.
\end{itemize}

The remainder of the paper is organized as follows. Section 2 outlines the related works including traditional counting methods, CNNs-based methods and column-based methods. Section 3 describes our proposed models. Section 4 shows the results of our experiments and section 5 concludes this paper.

\section{Related work}

\subsection{Traditional counting methods}

Traditional crowd counting algorithms are mainly divided into two categories: detection-based methods and regression-based methods.

Early researchers in crowd counting focus on detection-based methods. \citep{b9} used sliding window based detection algorithms to estimate the number of people in images. Also some low-level features such as histogram oriented gradients (HOG), Haar wavelets and edge were often extracted from human heads or human bodies for human detection \citep{b10,b11,b12}. While for partially occluded pedestrians, detection is disappointing.

Hence, regression-based methods are gradually used to solve the problem of crowd counting. Regression-based methods aim to learn the mapping function from low-level features in images such as foreground and texture to the count or density \citep{b13,b14,b15}. These regression methods are more efficient than detection methods, however, they do not fully utilized information in images.

\subsection{CNN-based counting methods}
Recently, CNN-based methods have demonstrated significant improvements over the traditional methods. Different network architectures are designed to handle various challenges such as background noise and scale changes \citep{b6,b8,b16,b33,b47,b61}. 

The crowd density we want to estimate is the number of people per unit area. However, the density maps will be severely affected by background noise, that is, the background terms with similar texture features to congested crowd scenes will be mistaken as heads easily. With attention model succeeded in various computer vision tasks, many researchers attempted to use the attention method to deal with background noise in crowd counting \citep{b3,b24,b26,b27,b39,b41,b42,b5,b8}. Some researchers use refinement-based algorithms to focus the crowd region and improve the quality of density maps \citep{b3,b24,b39}. \citep{b24} proposed a dual path multi-scale fusion network architecture to generate the final high-quality density maps by fusing attention maps and density maps. \citep{b3} devised a from-coarse-to-fine progressive attention mechanism to better focus on the crowd area for people count estimation. \citep{b39} introduced the Spatial-/Channel-wise Attention Models to alleviate the mistaken estimation for background regions. \citep{b2} proposed an attention-injective deformable convolutional network for crowd understanding that can suppress background noise in highly congested noise scenes. Another way to improve the performance of crowd counting is to adopt the idea of classification \citep{b26,b41,b42,b5}. \citep{b26} designed an attention module to adaptively estimate the crowd density based on its real density conditions with detection and regression. \citep{b41,b42} incorporated non-local attention mechanism to conquer huge scale variations. \citep{b5} provided different attention masks related to regions of different density levels aiming to attenuate the estimation errors in different regions. These attention methods could effectively reduce background interference.However, these attention methods use attention map as mask to do pixel-wise product, which may ignore the relationship between pixels.

\begin{figure*}[htbp]
\centering
\centerline{\includegraphics[scale=0.2]{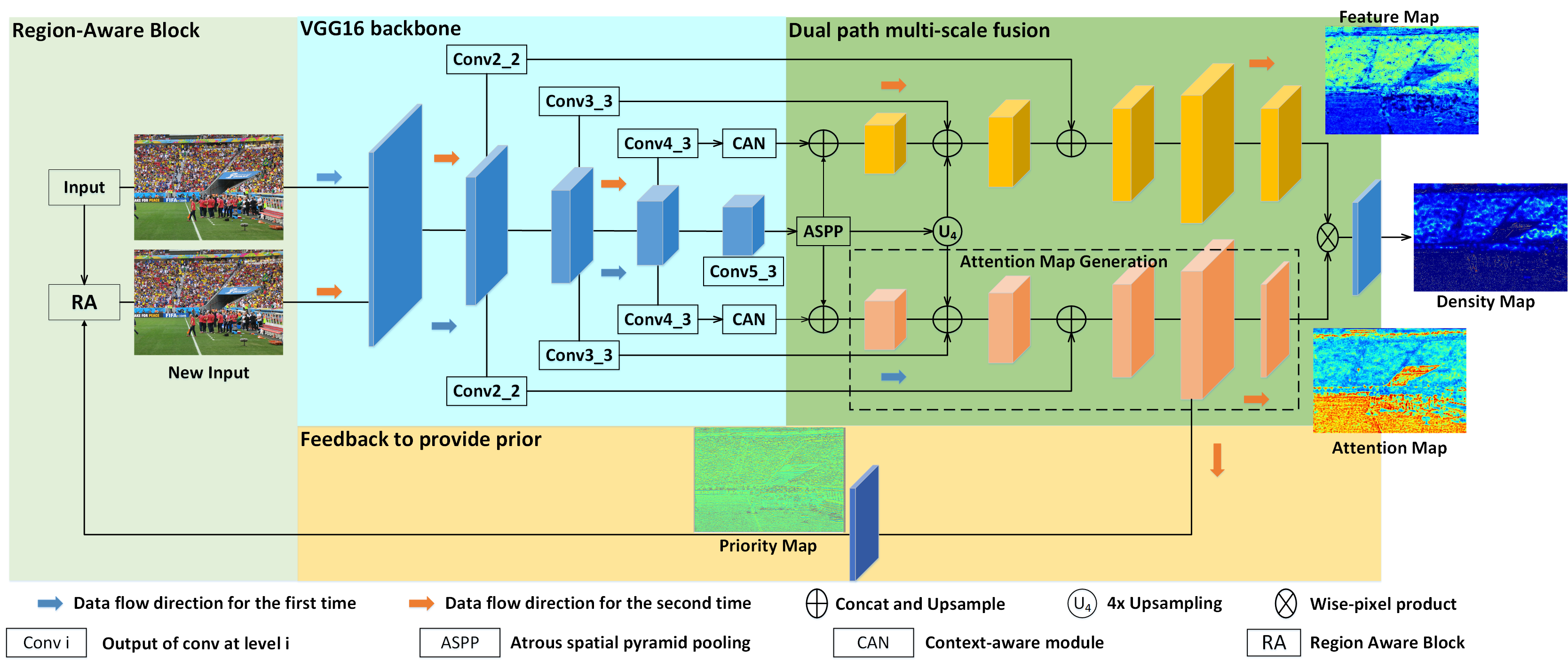}}
\caption{The architecture of the proposed region-aware network for crowd counting.}
\label{fig2}
\end{figure*}

Scale variation is also a problem that has been long recognized. Most works handled the large-scale variations issue using different architectures \citep{b18,b20,b23}.  \citep{b18} designed a Multi-Column Convolutional Neural Network (MCNN) architecture to estimate crowd number accurately in a single image from almost any perspective. Based on multi-scale CNN architecture, \citep{b20} designed the independent CNN regressor with different receptive fields and trained a switch classifier relay the crowd scene patch to the best CNN regressor. In contrast to these methods that propose specific architectures directly addressing scale variations, the recent methods concentrate on incorporating related information like high-level semantic information \citep{b17,b21} and contextual information \citep{b19,b22} respectively into the network. These related information is useful for network to understand the congested scenes. For example, \citep{b17} used a combination of high-level semantic information and the low-level features from deep learning framework to deal with large scale variations for estimating crowd density. \citep{b21} incorporated a high-level prior into the density estimation network enabling the network to learn globally relevant discriminative features for lower count error \citep{b21}. And \citep{b19} designed an end-to-end CNN architecture to predict both local and global count by making use of contextual information. \citep{b22} proposed an end-to-end trainable deep architecture that can adaptively encodes the scale of the contextual information aiming to accurately predict crowd density. These methods could alleviate the problem of scale variation to a certain extent. However, these methods may not capture sufficient global contextual information as they focus on local regions.

\subsection{Column-Based methods}
The patch-based and super pixel-based operations are common and consistent with human visual system. This is consistent with common cognition. However, Deep Convolutional Neural Networks (DCNNs) likely achieve an object recognition competence through a set of mechanisms that are distinct from those in humans \citep{b53}. Thus, the patch-based and super pixel-based operations may not be optimal choices for some applications. In the task of place recognition within 3D point cloud, Scan Context \citep{b57}, made a column-wise comparison to achieve effective localization for dynamic objects, by considering global information. In similar way, Kim proposed an analogous column scanning method named Scan Context Image, to improve the localization performance in SLAM task \citep{b58}. Image transformer flattens the input tensor in raster-scan order, and computes 1D local attention (similar to column-based operation) for generating natural-looking images \citep{b56}. Szeskin proposed a custom column based convolutional neural network, which is used in the classification of light scattering patterns in columns of vertical pixel-wide vectors in OCT slices \citep{b59}. 
These works use a similar idea of column-based operation and have achieved good results, which shows that column-based operation is reasonable in some works. Thus, we consider designing column-based region-aware block, which is used to adaptively encoder the global context information into features.

\section{Proposed method}

As discussed above, we aim to deal with the issues of background noise and scale variation. Human’s Top-Down visual perception mechanism can well deal with these issues. When human do crowd counting, according to the goal of the current task and prior knowledge, they firstly focus on crowd regions. Meanwhile, people would not trouble in scale variation as they will take the context information into consideration with a global receptive field. Inspired by this, we proposed a feedback structure with Region-Aware (RA) block modeling human’s Top-Down visual perception mechanism for crowd region enhancement and context information capturing through global receptive field. 

\subsection{Model architecture}

As shown in Fig. \ref{fig2}, the model architecture contains 4 components, VGG16 backbone, Dual path multi-scale fusion \citep{b28}, Feedback to provide prior and Region-Aware block. Input images are first fed into VGG16 backbone feature map extractor to extract multi-scale features. Then, the features of high-level semantics information are passed through context-aware module (CAN) \citep{b22}  and atrous spatial pyramid pooling (ASPP) \citep{b33} to obtain the scale-aware contextual features. CAN module combines features obtained in Conv4\_3 using multiple receptive field sizes of average pooling operation. The pooling output scales are 1,2,3,6. ASPP module applies dilated convolution with different rates (1,6,12,18) to features obtained in Conv5\_3 for multi-scale information. There is a skip connection between conv5\_3 and conv3\_3 for embedding high-level semantical information \citep{b28}. Then, the Dual path multi-scale fusion uses concatenate and up-sample to fuse these multi-scale features to generate priority map. The priority map would provide input images prior information that where are the important regions through feedback. Then, input images and corresponding priority map are put into the Region-Aware block together to obtain new inputs which would enhance the crowd regions and contain global context information. The new inputs are passed through the encoder-decoder based deep convolutional neural networks again to generate feature map and corresponding attention map. At last, the feature map and corresponding attention map would be fused to generate the final high-resolution density map.

\subsection{Feedback to provide prior}

According to goal of current task and prior knowledge, human could focus on region of interest. With similar idea, we would like to boost the crowd regions in input images for reducing background interference. Given a set of $N$ training images $ {\left\{ {{Q_i}} \right\}_{1 \le i \le N}} $, our goal is to train corresponding priority maps $ {\left\{ {{A_i}} \right\}_{1 \le i \le N}} $ to focus on crowd regions in input images. In order to combine more efficient information, we choose the output fusing the multi-scale features which generated from the VGG16 backbone as shown in Fig. \ref{fig2}:

\begin{equation}
{f_i} = {F_{vgg}}\left( Q \right),i = \left\{ {2,3,4,5} \right\}
\end{equation}

\begin{equation}
{A_i} = {F_{amg}}\left( {{f_2},{f_3},{f_4},{f_5}} \right)
\end{equation}

\noindent where ${F_{vgg}}$  is the VGG16 backbone that extracts multi-scale features  ${f_i}$; ${F_{amg}}$ is the decoder network that fuses multi-scale features by using bilinear interpolation and concatenation.

\begin{figure}[htbp]
\centering
\subfigure[Input image]{
\includegraphics[width=3.5cm,height=2.8cm]{fig3_a.png}
\label{fig3(a)}
}
\quad
\centering
\subfigure[Priority map]{
\includegraphics[width=3.5cm,height=2.8cm]{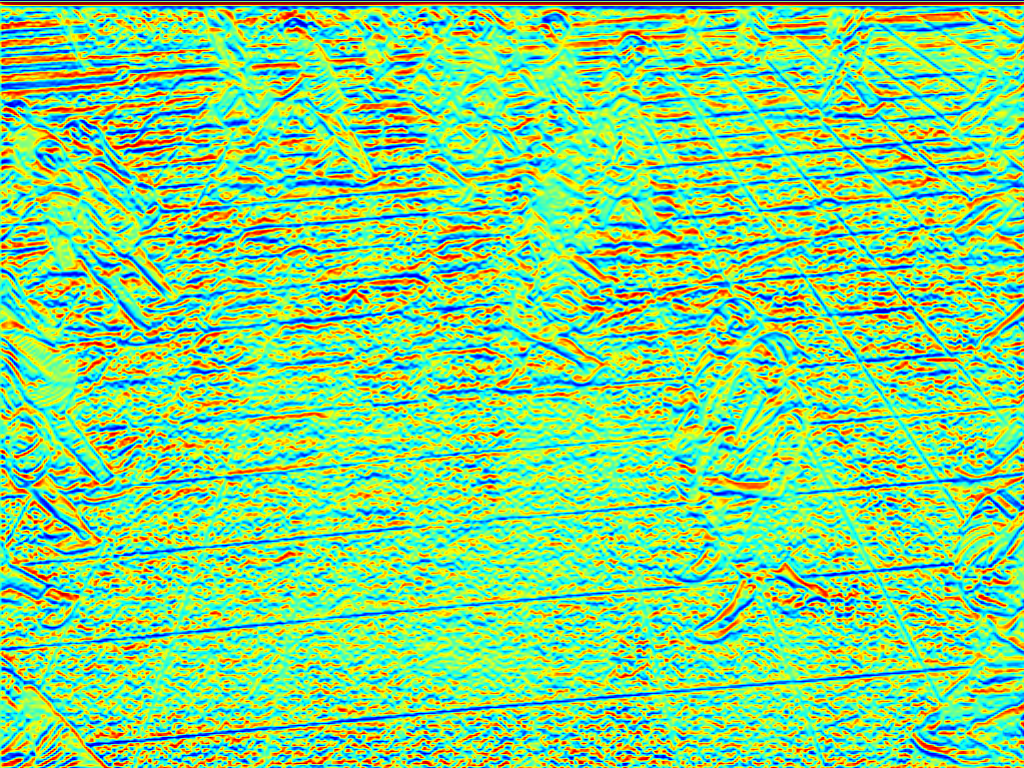}
\label{fig3(b)}
}
\quad
\centering
\subfigure[New input]{
\includegraphics[width=3.5cm,height=2.8cm]{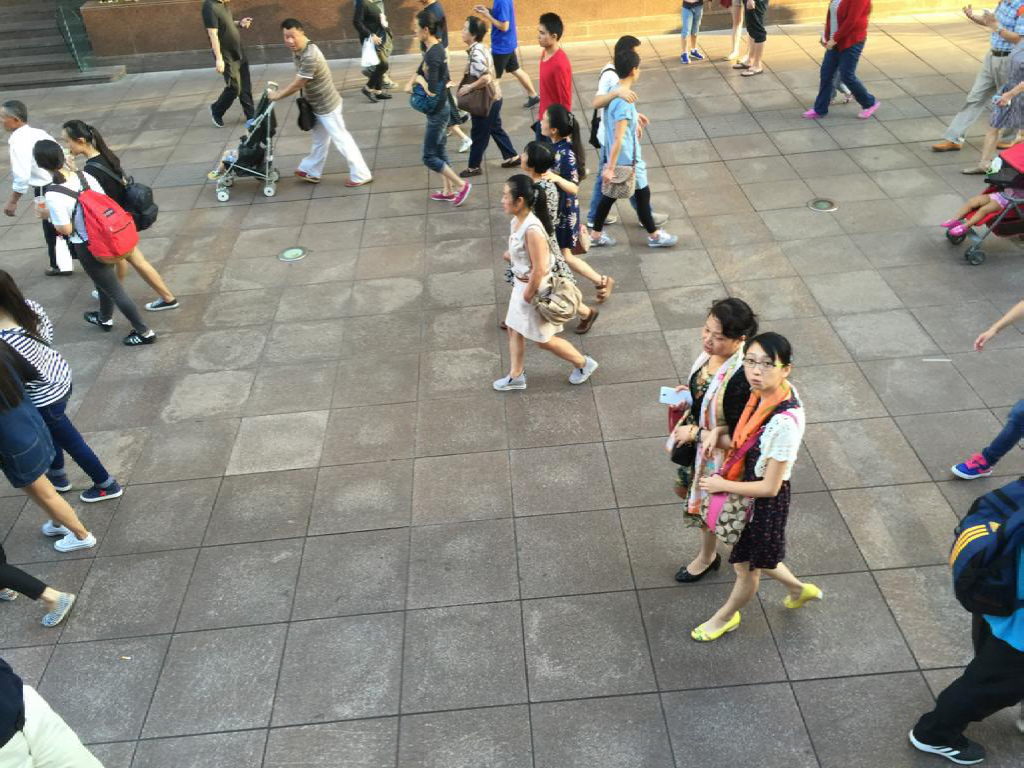}
\label{fig3(c)}
}
\quad
\centering
\subfigure[Difference between (a)  Input images and (c) New inputs]{
\includegraphics[width=3.5cm,height=2.8cm]{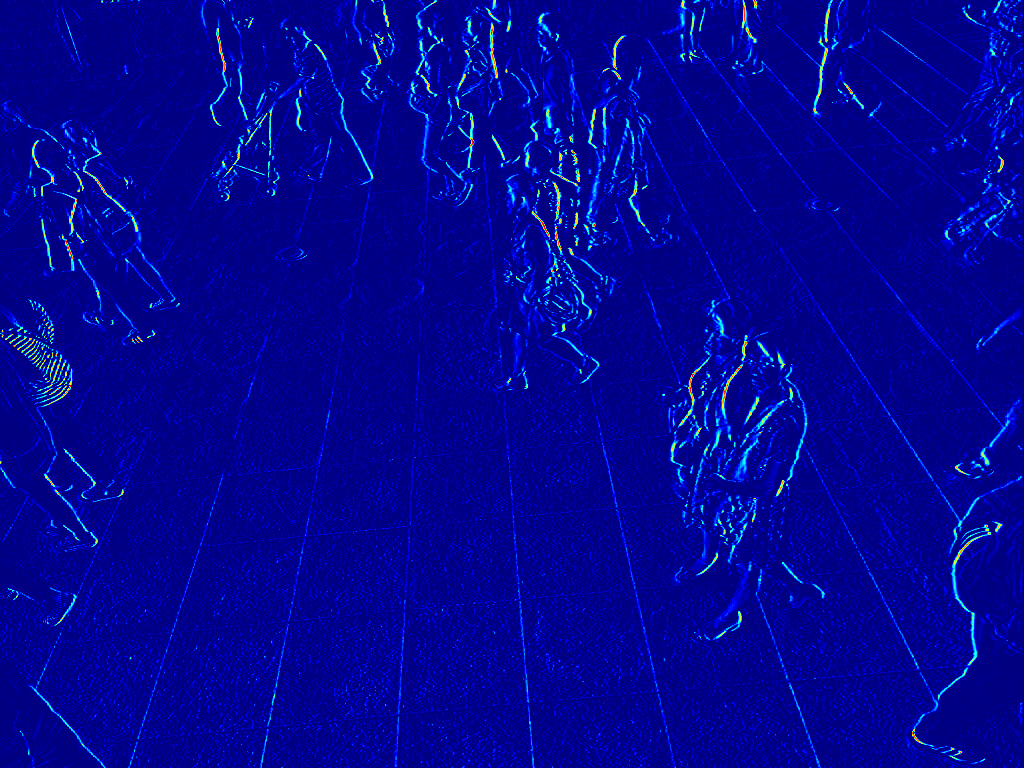}
\label{fig3(d)}
}
\quad
\caption{ Visualization of the outputs of RANet. (a) is input image; (b) is the corresponding priority map; (c) is the new input obtained from the Region-Aware Block; (d) is obtained by subtracting (a) input images from (c) new inputs.}
\label{fig3}
\end{figure}

The priority map could find the boundary between persons and background as Fig. \ref{fig3(b)} shows. Taking the priority map to boost input images will provide prior about candidate crowd region. The prior enables the inputs pay more attention to those crowd regions. Traditional attention map is an image-sized weight map where crowd regions have higher values \citep{b2}. They often take an element-wise multiplication while ignore the context information in the crowd region. Different from the general attention mechanism, we enhance the input images by learning the similarity between input images and priority maps and embedding the obtained relevance into inputs as Fig. \ref{fig4} shows. The details will be illustrated in the following Region-Aware Block.

\subsection{Region-Aware Block}

\begin{figure}[htbp]
\centering
\centerline{\includegraphics[scale=0.5]{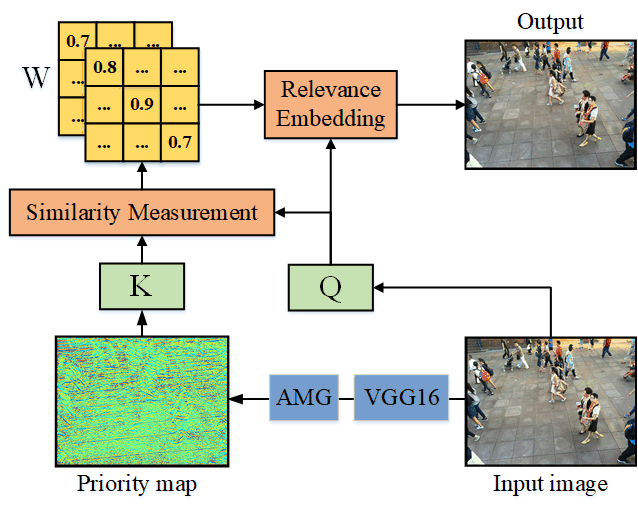}}
\caption{The proposed region-aware block.}
\label{fig4}
\end{figure}

As the heart of convolution neural networks, standard convolutions always exploit the same filters and pooling operations over the whole image. This means that standard convolutions can only capture the local spatial correlation (except the ones at top-most layers). Lack of global receptive field is hard to account for large scale variation and capture context information fully.

In order to deal with scale variation, we introduce global context information into the input images by designing the Region-Aware Block. Specifically, as Fig. \ref{fig4} shows, we scan the whole input images and its priority maps in the form of column vector to estimate their similarity for ontaining relevance matrix. Then we embed the relevance matrix $W$ into the input images to build global relationships between pixels. By doing this, the network can encoder context information for understanding scale variation and expand the size of receptive field.

\subsubsection{Similarity Measurement: Global Information Capturing}

Similarity Measurement aims to obtain global relationships between pixels in input images by estimating the similarity between each column in the input images and that in its priority maps.
 
Traditional attention mechanism takes pixel-wise product between the input image and its attention map. However, such a pixel-wise product may cause lack of context information and local receptive field as they focus on a single pixel and ignore the relationship between pixels. This would not account for scale variation well in images.

In order to capture global relationships between pixels, we scan the whole input images and its priority maps in the form of column vector to estimate their similarity. That is, we estimate the similarity between each column vector in input images and that in its priority maps to obtain weight matrix representing relevance between pixels. 

More specifically, the sizes of input image $Q$ and its priority map $A$ are both $N \times M$, denoted as:

\begin{equation}
{\rm{Q = }}\left[ {\begin{array}{*{20}{c}}
{{q_{11}}}& \cdots &{{q_{1m}}}\\
 \vdots & \ddots & \vdots \\
{{q_{n1}}}& \cdots &{{q_{nm}}}
\end{array}} \right]
\end{equation}

\begin{equation}
{\rm{A = }}\left[ {\begin{array}{*{20}{c}}
{{a_{11}}}& \cdots &{{a_{1m}}}\\
 \vdots & \ddots & \vdots \\
{{a_{n1}}}& \cdots &{{a_{nm}}}
\end{array}} \right]
\end{equation}

Then for each column $Q\left( {:,i} \right)$ in Q and $A\left( {:,j} \right)$ in $A$, we calculate the similarity ${s_{i,j}}$ between these two columns by inner product:

\begin{equation}
{s_{i,j}} = \left\langle {Q\left( {:,i} \right)\left. {,A\left( {:,j} \right)} \right\rangle } \right. = \sum\limits_{r = 1}^n {{q_{ri}}{a_{rj}}} 
\end{equation}

\begin{equation}
S = {Q^T} \times A = \left[ {\begin{array}{*{20}{c}}
{{s_{11}}}& \cdots &{{s_{1m}}}\\
 \vdots & \ddots & \vdots \\
{{s_{m1}}}& \cdots &{{s_{mm}}}
\end{array}} \right]
\end{equation}

Then, we use softmax function to get the relevance matrix W from similarity matrix $S$.

\begin{equation}
\begin{aligned}
W =& soft\max(S) = soft\max({Q^T} \times A) 
\\ =& \left[ {\begin{array}{*{20}{c}}
{{w_{11}}}& \cdots &{{w_{1m}}}\\
 \vdots & \ddots & \vdots \\
{{w_{m1}}}& \cdots &{{w_{mm}}}
\end{array}} \right]
\end{aligned}
\end{equation}

The value of ${w_{ij}}$ can be regarded as an index, which represents the relevance between $i - th$ column in input image and $j - th$ column in corresponding priority map. Consequently, the relevance matrix W can provide input image with access to global information. In other word, the relevance matrix W is further embedded into input images to adaptively build relationship between pixels..

\subsubsection{Relevance Embedding: Adaptive Recalibration}

We proposed a relevance embedding module to exploit relevance matrix obtained from similarity measurement to build global relationships between pixels. That is, we use the similarity between column vectors in input image and priorty map to adaptively recalibrate relationship between pixels.

More specifically, we calculate the output $O$ by inner product between the input image $Q$ and relevance matrix $W$ as follows:

\begin{equation}
O = Q \times {W^T} = \left[ {\begin{array}{*{20}{c}}
{{o_{11}}}& \cdots &{{o_{1m}}}\\
 \vdots & \ddots & \vdots \\
{{o_{n1}}}& \cdots &{{o_{nm}}}
\end{array}} \right]
\end{equation}

\begin{equation}
\begin{array}{l}
{o_{ij}} = \sum\limits_{r = 1}^m {{q_{ir}}{w_{jr}}} {\rm{ = }}\sum\limits_{r = 1}^m {{q_{ir}} \left[soft\max \left( {\sum\limits_{l = 1}^n {{q_{lj}} \cdot {k_{lr}}} } \right)\right]} 
\end{array}
\end{equation}

Discussion: we can build global relationship between pixels through similarity measurement and relevance embedding. The new input is displayed in Fig. \ref{fig3(c)}. Compare Fig. \ref{fig3(c)} and Fig. \ref{fig3(d)}, we could find that the new input could enhance the crowd regions. 

Afterward, we train the encoder-decoder networks based deep convolutional neural networks again using the output as new input.

\subsection{Loss Function}

As we introduce the priors and context information into input images, we use the Bayesian loss as the loss function \citep{b34}. Suppose that $x\left( {{x_m} = m:m = 1,2,...,M} \right)$ is a random variable that represents the spatial location and $y\left( {{y_n} = n:n = 1,2,...,N} \right)$ is a random variable that denotes the annotated head points, ${y_0}$ is the background pixels. According to bayes’ theorem, the posterior probability of  ${x_m}$ obtaining the annotation ${y_n}$ and background label ${y_0}$ can be calculated as:

\begin{equation}
p\left( {{x_m}|{y_n}} \right) = \frac{1}{{\sqrt {2\pi } \delta }}\exp \left( { - \frac{{\left\| {{x_m} - {y_n}} \right\|_2^2}}{{2{\delta ^2}}}} \right)
\end{equation}

\begin{equation}
p\left( {{x_m}|{y_0}} \right) = \frac{1}{{\sqrt {2\pi } \delta }}\exp \left( { - \frac{{{{\left( {d - {{\left\| {{x_m} - y_n^m} \right\|}_2}} \right)}^2}}}{{2{\delta ^2}}}} \right)
\end{equation}

\begin{equation}
\begin{aligned}
 p\left(y_{n} \mid \mathbf{x}_{m}\right) &=\frac{p\left(\mathbf{x}_{m} \mid y_{n}\right) p\left(y_{n}\right)}{\sum_{n=1}^{N} p\left(\mathbf{x}_{m} \mid y_{n}\right) p\left(y_{n}\right)+p\left(\mathbf{x}_{m} \mid y_{0}\right) p\left(y_{0}\right)} 
\\
 &=\frac{p\left(\mathbf{x}_{m} \mid y_{n}\right)}{\sum_{n=1}^{N} p\left(\mathbf{x}_{m} \mid y_{n}\right)+p\left(\mathbf{x}_{m} \mid y_{0}\right)} 
\end{aligned}
\end{equation}

\noindent where $\delta $ is the variance of a 2D Guassian distribution, $y_n^m$ denotes the nearest head point of ${x_m}$ and $d$ is a parameter that defines the background points by controlling the distance between the head and background points. Due to the different density of scenes in different datasets, the choice of parameter $d$ is also different which will be shown in Table \ref{tab1}.

The last equation is simplified with the assumption $p\left( {{y_n}} \right) = p\left( {{y_0}} \right) = 1/\left( {N + 1} \right)$. Then the estimated counts for each person and background are defined as: 

\begin{equation}
\begin{array}{l}
E\left[c_{n}\right]=\sum_{m=1}^{M} p\left(y_{n} \mid \mathbf{x}_{m}\right) \mathbf{D}^{e s t}\left(\mathbf{x}_{m}\right) \\
E\left[c_{0}\right]=\sum_{m=1}^{M} p\left(y_{0} \mid \mathbf{x}_{m}\right) \mathbf{D}^{e s t}\left(\mathbf{x}_{m}\right)
\end{array}
\end{equation}

\noindent where $\mathbf{D}^{e s t}\left(\mathbf{x}_{m}\right)$ is the estimated density map and $\mathbf{x}_{m}$ denotes a 2D pixel location.

In this case, we would like the foreground count at each annotation point equals to one and the background count to be zero. Thus, the final loss function is as follows:

\begin{equation}
\mathcal{L}^{\text {Bayes }+}=\sum_{n=1}^{N} \mathcal{F}\left(1-E\left[c_{n}\right]\right)+\mathcal{F}\left(0-E\left[c_{0}\right]\right)
\end{equation}

\section{Experiment}
In this section, we present the experimental details and evaluation results on 4 public challenging datasets: ShanghaiTech \citep{b18}, UCF\_CC\_50 \citep{b36}, UCF-QRNF \citep{b35},  JHU-CROWD++ \citep{b37} and NWPU \citep{b49}.

\subsection{Evaluation Metrics and Implementation Details}
We use the following metrics mean absolute error (MAE), mean square error (MSE) and
mean normalized absolute error (NAE) to evaluate the performance of our method.

\begin{equation}
M A E=\frac{1}{N} \sum_{i=1}^{N}\left|C_{i}-C_{i}^{G T}\right|
\end{equation}

\begin{equation}
M S E=\sqrt{\frac{1}{N} \sum_{i=1}^{N}\left|C_{i}-C_{i}^{G T}\right|^{2}}
\end{equation}

\begin{equation}
N A E=\frac{1}{N} \sum_{i=1}^{N} \frac{\left|C_{i}-C_{i}^{G T}\right|}{C_{i}^{G T}}
\end{equation}

\noindent where $N$ is the number of test images, $C_{i}$ and $C_{i}^{G T}$ are the estimated count and the ground truth respectively.

We do the image augmentation using random crop. As shown in Table \ref{tab1}, the sizes of the cropped images differ across datasets. Therefore, when training different datasets, we use different learning rates and batch sizes.

\renewcommand\arraystretch{0.92}
\begin{table*}[htbp]
\caption{Estimation errors on ShanghaiTech dataset}
\begin{center}
\begin{tabular}{lllll}
\hline Method & Part A & & Part B & \\
& MAE & MSE & MAE & MSE \\
\hline MCNN \citep{b18} & $110.2$ & $173.2$ & $26.4$ & $41.3$ \\
Switch-CNN \citep{b20} & $90.4$ & $135.0$ & $21.6$ & $33.4$ \\
CSR-Net \citep{b44} & $68.2$ & $115.0$ & $10.6$ & $16.0$ \\
SA-Net \citep{b45} & $67.0$ & $104.2$ & $8.4$ & $13.6$ \\
CAN \citep{b22} & $62.3$ & 100 & $7.8$ & $12.2$ \\
MBTTBF \citep{b46} & $60.2$ & $94.1$ & $8.0$ & $15.5$ \\
ADCcrowdNet \citep{b2} & $70.9$ & $115.2$ & $7.7$ & $12.9$ \\
LSC-CNN \citep{b47} & $66.4$ & $117.0$ & $8.1$ & $12.7$ \\
BL \citep{b34} & $62.8$ & $101.8$ & $7.7$ & $12.7$ \\
SFCN \citep{b48} & $67.0$ & $104.5$ & $8.4$ & $13.6$ \\
CG-DRCN-CC-Res101 \citep{b37} & ${60.2}$ & $\textbf{94.0}$ & $7.5$ & $12.1$ \\
M-SFANet \citep{b28} & $62.49$ & $106.11$ & $\textbf{6.38}$ &  $\textbf{10.22}$ \\
OURS & $\textbf{57.92}$ & $99.23$ & $7.15$ & $11.86$ \\
\hline
\end{tabular}
\label{tab2}
\end{center}
\end{table*}

\begin{table}[htbp]
\caption{training settings for each datasets}
\begin{center}
\setlength{\tabcolsep}{0.7mm}{
\begin{tabular}{lcccc}
\hline \multicolumn{1}{c} {\textbf{ Dataset} } &\textbf{Learning rate} & \textbf{Batch size} & \textbf{Crop size} & $\mathrm{\textbf{d}}$ \\
\hline ShanghaiTech PartA & $1 \mathrm{e}-6$ & 8 & $256 \times 256$ & $0.1$ \\
ShanghaiTech PartB & $1 \mathrm{e}-6$ & 8 & $400 \times 400$ & $0.1$ \\
UCF\_CC\_50 & $5 \mathrm{e}-4$ & 5 & $512 \times 512$ & $0.15$ \\
UCF-QNRF & $5 \mathrm{e}-4$ & 5 & $512 \times 512$ & $0.15$ \\
JHU-CROWD & $5 \mathrm{e}-4$ & 5 & $512 \times 512$ & $0.15$ \\
NWPU & $5 \mathrm{e}-4$ & 5 & $512 \times 512$ & $0.15$ \\
\hline 
\end{tabular}}
\label{tab1}
\end{center}Note: "d" is the parameter that described in the loss function.
\end{table}

\subsection{ShanghaiTech dataset}
ShanghaiTech \citep{b18} crowd counting dataset contains 1198 labeled images with 330165 people annotated totally. The dataset is divided into two parts named Part A and Part B. Part A consists of 482 (300 for train, 182 for test) images with highly congested scenes collected from the internet. The images in Part A are highly dense with crowd counts between 33 to 3139. While Part B contains 716 (400 for train, 316 for test) images taken from busy streets in Shanghai. The images in Part B are less dense with the number of people varying from 9 to 578. Because of limited numbers of training samples, we pre-train our models on UCF-QNRF.

We evaluate our model and compare it to other 12 recent methods in Table \ref{tab2}. The results in Table \ref{tab2} show that our model can achieve better performance with 7.31\% MAE and 6.48\% MSE improvement compared with base model(M-SFANet) on Part A. It can also be observed that the proposed method is able to achieve comparable performance with the base model (M-SFANet) on Part B. The effect of our model is similar to that of other models in ShanghaiTech dataset. The reason may be that ShanghaiTech dataset contains less people than other popular datasets and scale variation are not so dramatic compared to other challenging datasets.

\begin{figure}[htbp]
\centering
\centerline{\includegraphics[scale=0.35]{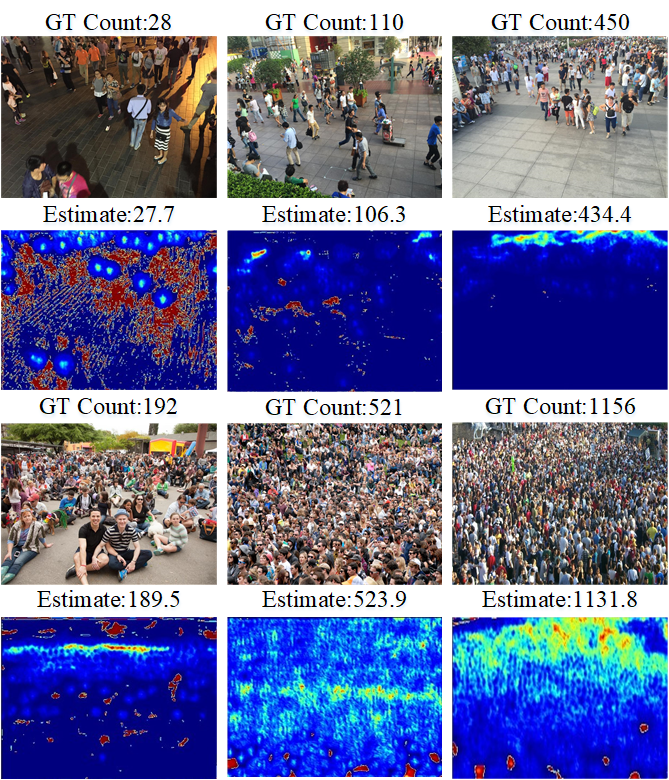}}
\caption{Visualization of input images from ShanghaiTech and corresponding estimated density maps. The first and third rows are samples images from ShanghaiTech Part B and ShanghaiTech Part A, respectively. The second and fourth rows are the corresponding estimated density maps from our proposed method.}
\label{fig5}
\end{figure}

To further explore the reasons for the different performance of our method on the two datasets, we select representative images from the two datasets and compare the estimated density maps in Fig. \ref{fig5}. Compared to Part A, the scenes in Part B are more simple and monotonous. More specifically, the scenes of Part B are mainly composed of busy streets and consist of less people. Therefore, the issues of background noise and scale variation are not prominent in Part B dataset. This makes our method still performs honorably but looses its edge compared to the others in Part B dataset. On the contrary, in Part A dataset which is collected from the internet including various diverse scenarios, our model performs better. This further illustrates the superiority and robustness of our approach under a variety of diverse scenarios.

\subsection{UCF\_CC\_50 dataset}

UCF\_CC\_50 \citep{b36} is an extremely dense crowd dataset including 50 images of different resolutions. The numbers of head annotations range from 94 to 4543 with an average number of 1280. To better evaluate model performance, 5-fold cross-validation is performed following the standard setting in \citep{b36}. Because of limited numbers of training samples, we pre-train our models on UCF-QNRF to speed up convergence.

Table \ref{tab3} shows the results on UCF\_CC\_50 dataset. The proposed method is compared with other recent works. It can be observed that our model obtains the best performance with 4.51\% MAE and 20.7\% MSE improvement compared with the second best approach M-SFANet.

\begin{figure}[htbp]
\centering
\subfigure
{
\includegraphics[width=2.2cm,height=1.8cm]{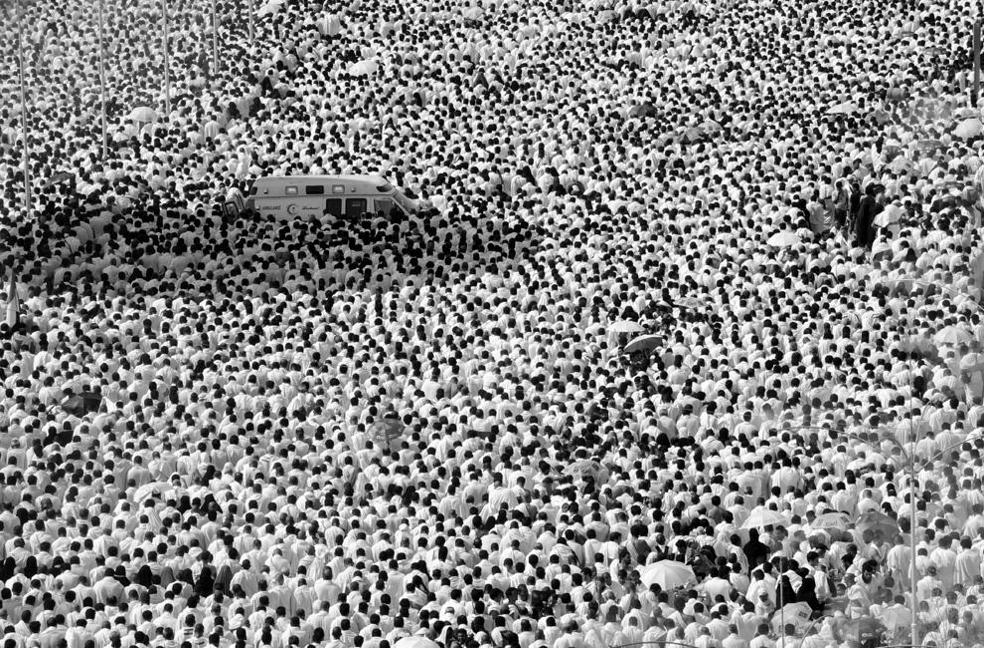} 
}
\quad
\centering
\subfigure{
\includegraphics[width=2.2cm,height=1.8cm]{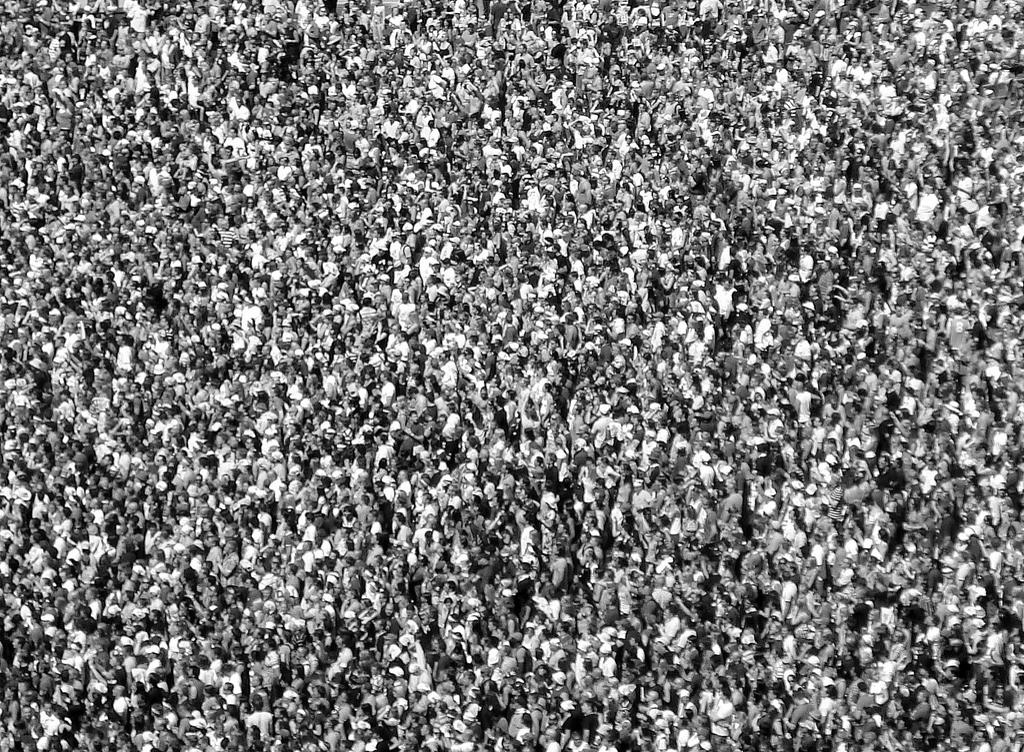}
}
\quad
\centering
\subfigure{
\includegraphics[width=2.2cm,height=1.8cm]{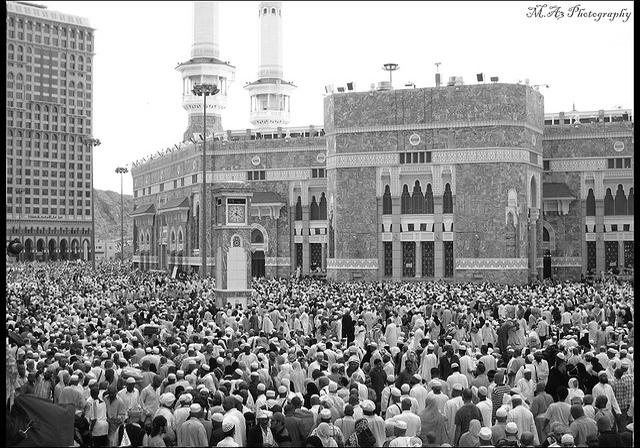}
}
\caption{Some representative sample images from UCF\_CC\_50 datasets. These sample images are all extremely dense crowd images. More precisely, images in this dataset are almost extremely dense scenes.}
\label{fig6}
\end{figure}

On the UCF\_CC\_50 dataset, we consistently and clearly outperform all other methods. As shown in Fig. \ref{fig6}, images in UCF\_CC\_50 dataset are mostly extremely dense crowd images. This makes context more informative and our approach state-of-the-art. What’s more, compared to MAE, the improvement of MSE is more remarkable. Compared with MAE, MSE assesses the estimated deviation of the overall data. The smaller the MSE, the more accurate our estimation of the number of people in each image. This could further prove the robustness of our method. In other word, our approach can evaluate approximate number close to ground truth in mostly extremely dense crowd scenes. This is consistent with the behavior of humans observing dense scenes, aiming to get the approximate number of people in dense scenes.

\renewcommand\arraystretch{0.92}
\begin{table}[htbp]
\caption{Estimation errors on UCF\_CC\_50 dataset}
\begin{center}
\begin{tabular}{ccc}
\hline Method & MAE & MSE \\
\hline MCNN \citep{b18} & $377.6$ & $509.1$ \\
Switch-CNN \citep{b20} & $318.1$ & $439.2$ \\
CSR-Net \citep{b44} & $266.1$ & $397.5$ \\
SA-Net \citep{b45} & $258.4$ & $334.9$ \\
CAN \citep{b22} & 107 & 183 \\
MBTTBF \citep{b46} & $233.1$ & $300.9$ \\
ADCcrowdNet \citep{b2} & $273.6$ & $362.0$ \\
LSC-CNN \citep{b47} & $225.6$ & $302.7$ \\
BL \citep{b34} & $229.3$ & $308.2$ \\
SFCN \citep{b48} & $258.4$ & $334.9$ \\
M-SFANet \citep{b28} & $162.33$ & $276.76$ \\
OURS & $\textbf{155.01}$ & $\textbf{219.45}$ \\
\hline
\end{tabular}
\label{tab3}
\end{center}
\end{table}

\subsection{UCF-QNRF dataset}

UCF-QNRF \citep{b35} is a large and challenging dataset due to the extremely congested scenes. The dataset contains 1535 (1201 for train, 334 for test) jpeg images with 1251642 people in them. What’s more, it has a wide range of counts, complex environment and image resolutions. We train our model on UCF-QNRF with VGG-16bn pre-trained weights.

We evaluate our model and compare it to other recent works and results in Table \ref{tab4}. The results in Table \ref{tab4} indicate that our model can achieve better performance with 2.59\% MAE and 6.24\% MSE improvement compared with the second best approach M-SFANet. 

\begin{figure*}[htbp]
\centering
\centerline{\includegraphics[scale=0.42]{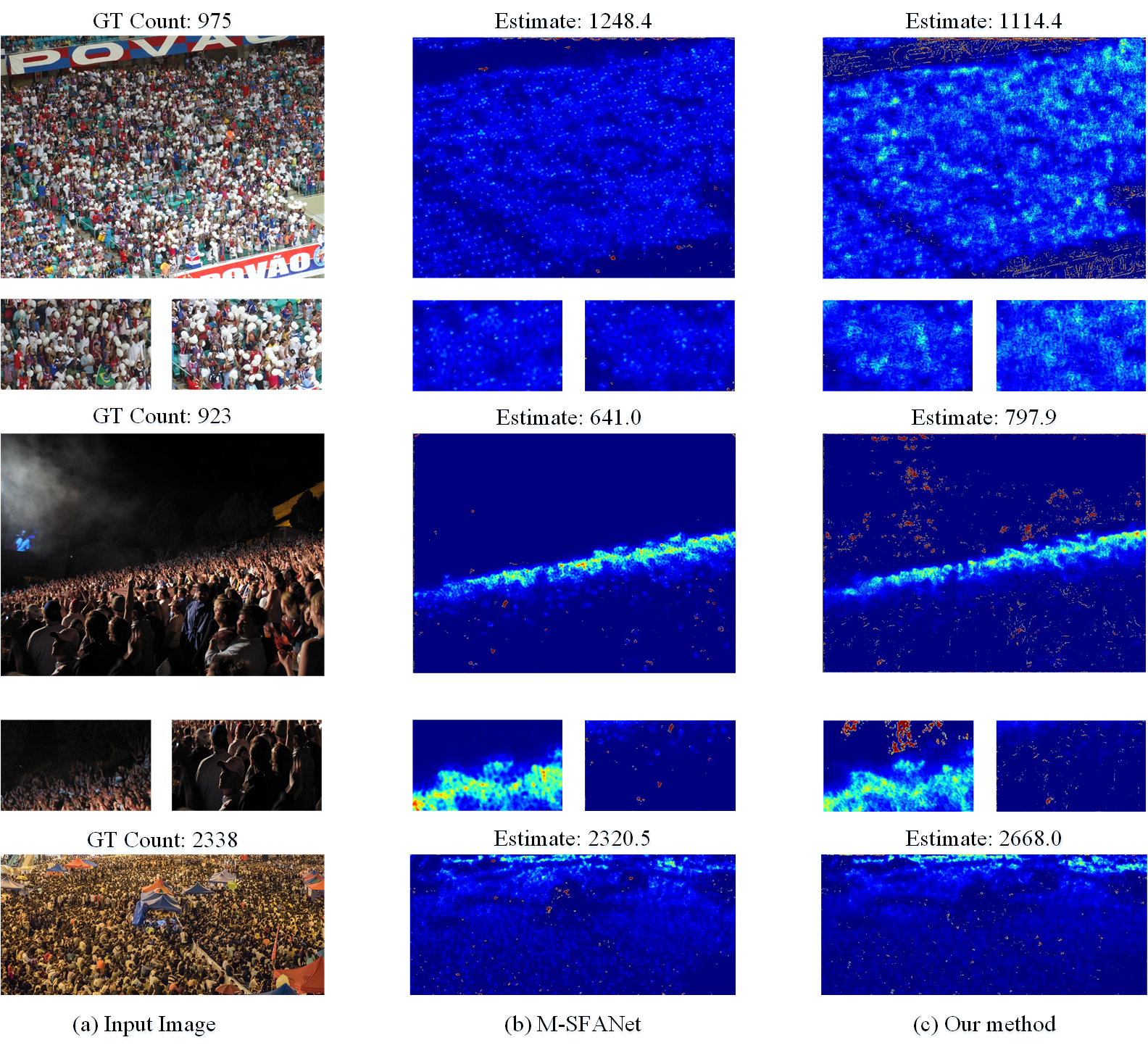}}
\caption{Input images form UCF-QNRF, along with the density maps generated by the M-SFANet and our proposed method. From left to right: The left column are input images; the middle column are density maps generated by M-SFANet and the right column are density maps generated by our proposed method. Compared to accurately localizing each person in M-SFANet, our method pays more attention to the crowd regions. What’s more, our method can obtain a larger received field for crowd regions.}
\label{fig7}
\end{figure*}

In Fig. \ref{fig7}, we show input images form UCF-QNRF, along with the density maps generated by the M-SFANet and our proposed method. 
Compared to accurately localizing each person in M-SFANet model, our method pays more attention to the crowd regions. This means that our method could take context information into consideration for more precise crowd counting. This is exactly in line with human’s Top-Down visual system: human scan a crowd image and instantly know the approximate number of human through overall perception of congestion degree of crowd regions. What’s more, as Fig. \ref{fig7} shows, our method can obtain a larger received field which is more closely match the distribution of the crowd regions.

\renewcommand\arraystretch{0.92}
\begin{table}[htbp]
\caption{Estimation errors on UCF-QNRF dataset}
\begin{center}
\begin{tabular}{lll}
\hline Method & MAE & MSE \\
\hline MCNN \citep{b18} & $277.0$ & $126.0$ \\
Switch-CNN \citep{b20} & 228 & 445 \\
CSR-Net\citep{b44} & $120.3$ & $208.5$ \\
CAN \citep{b22} & $212.2$ & $243.7$ \\
MBTTBF \citep{b46} & $97.5$ & $165.2$ \\
LSC-CNN \citep{b47} & $120.5$ & $218.2$ \\
BL \citep{b34} & $88.7$ & $154.8$ \\
SFCN \citep{b48} & $102.0$ & $171.4$ \\
CG-DRCN-CC-Res101 \citep{b37} & $95.5$ & $164.3$ \\
M-SFANet \citep{b28} & $85.6$ & $151.23$ \\
OURS & $\textbf{83.38}$ & $\textbf{141.79}$ \\
\hline
\end{tabular}
\end{center}
\label{tab4}
\end{table}

\subsection{JHU-CROWD++ dataset}

\begin{table}[htbp]
\caption{Estimation errors on JHU-CROWD++ dataset}
\begin{center}
\begin{tabular}{lll}
\hline Method & MAE & MSE \\
\hline MCNN \citep{b18} & $188.9$ & $483.4$ \\
CSR-Net\citep{b44} & $85.9$ & $309.2$ \\
SA-Net \citep{b45} & $91.1$ & $320.4$ \\
MBTTBF \citep{b46} & $81.8$ & $299.1$ \\
LSC-CNN \citep{b47} & $112.7$ & $454.4$ \\
SFCN \citep{b48} & $82.3$ & $328.0$ \\
CG-DRCN-CC-Res101 \citep{b37} & $71.0$ & $278.6$ \\
OURS & $\textbf{59.36}$ & $\textbf{257.56}$ \\
\hline
\end{tabular}
\end{center}
\label{tab5}
\end{table}

JHU-CROWD++ \citep{b37} contains 4372 images containing a total of 1.51 million dot annotations with an average of 346 dots per image and a maximum of 25000 dots. In comparison to most datasets, the JHU-CROWD++ dataset is a large dataset collecting under a variety of diverse scenarios and environment conditions. 

We evaluate our model and compare it to other recent works and results in Table \ref{tab5}. The results in Table \ref{tab5} show that our model can achieve better performance with 16.39\% MAE and 7.55\% MSE improvement compared with the second best approach CG-DRCN-CC-Res101.

\begin{figure}[htbp]
\centering
\centerline{\includegraphics[scale=0.35]{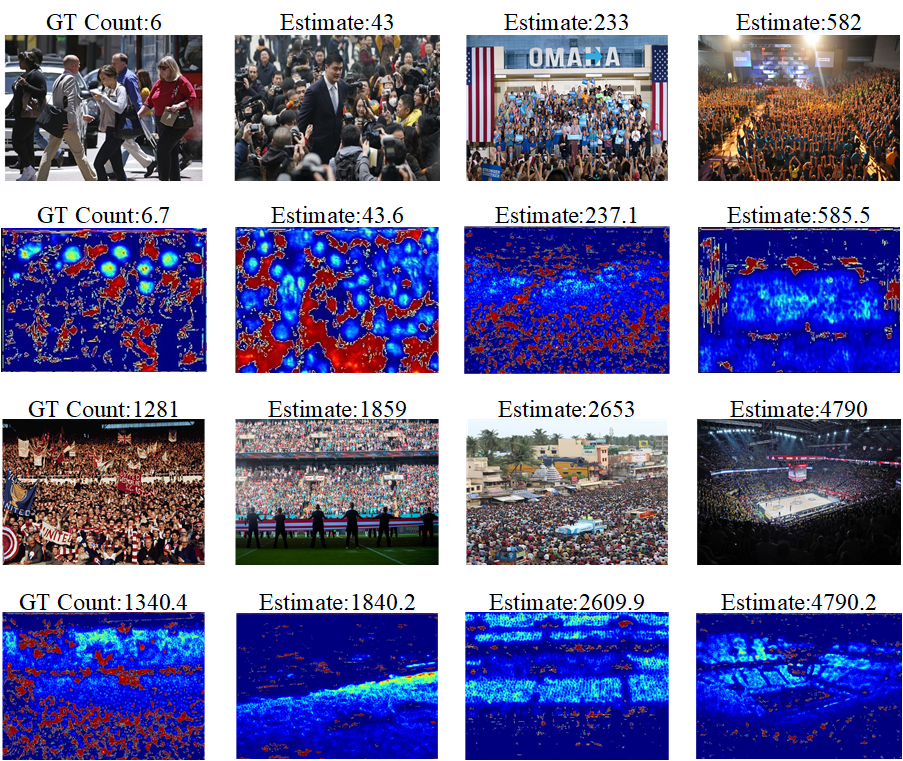}}
\caption{Visualization of estimated density maps from JHU-CROWD++. From left to right: the number of human gradually increases.}
\label{fig8}
\end{figure}

In Fig. \ref{fig8}, we show input images form JHU-CROWD++, along with the density maps generated by our proposed method. We can find that our proposed method has a good ability to estimate number in different dense scenarios. In other word, through introducing global context information into the input images by designing the Region-Aware Block, our approach can deal with scale variation well.

\subsection{NWPU dataset}
The NWPU dataset is the largest-scale and most challenging crowd counting dataset publicly available \citep{b49}. It is a large-scale congested crowd counting dataset that consists of 5,109 images crawled from the Internet, elaborately annotating 2,133,375 instances. The ground truth for test images set are not released and researchers could submit their results online for evaluation.

We evaluate our model and compare it to other recent works and results in Table \ref{tab11}. The results in Table \ref{tab11} show that our model significantly outperforms the state-of-the-art methods. Notably, on the NWPU test (obtained by submitting to the evaluation server), our model reduces the MAE and MSE by a large margin, from 88.4 to 77.5 in MAE and from 388.6 to 356.8 in MSE.

\begin{table*}[htbp]
\caption{Comparison with state-of-the-art methods on NWPU validation and test sets.}
\begin{center}
\begin{tabular}{cccccc}
\hline Method & \multicolumn{2}{c}{ Validation set } & \multicolumn{3}{c}{ Test set } \\
& MAE & MSE & MAE & MSE & NAE \\
\hline MCNN \citep{b18} & $218.5$ & $700.6$ & $232.5$ & $714.6$ & $1.063$ \\
CSRNet\citep{b44} & $104.8$ & $433.4$ & $121.3$ & $387.8$ & $0.604$ \\
CAN\citep{b22} & $93.5$ & $489.9$ & $106.3$ & $386.5$ & $0.295$ \\
BL\citep{b34} & $93.6$ & $470.3$ & $105.4$ & $454.2$ & $0.203$ \\
SFCN\citep{b48} & $95.4$ & $608.3$ & $105.7$ & $424.1$ & $0.254$ \\
DM-Count\citep{b50} & $70.5$ & $\textbf{357.6}$ & $88.4$ & $388.6$ & $\textbf{0.169}$ \\
OURS & $\textbf{65.3}$ & $432.9$ & $\textbf{77.5}$ & $\textbf{365.8}$ & $0.228$ \\
\hline
\end{tabular}
\end{center}
\label{tab11}
\end{table*}

\subsection{Ablation Study}

\subsubsection{complexity of the model}
  To evaluate the complexity of our method, we have conducted ablation study on NWPU dataset in Table \ref{tab6}. To exclude interference from other factors, we conducted the experiment on the same experimental environment, and reported the results in the largest online benchmark NWPU \citep{b49}. As shown in Table \ref{tab6}, our model does not have an advantage in model parameters and inference speed. However, our model has achieved better performance in the crowd counting. Moreover, our model could also achieve real-time crowd counting at a speed of 0.095 seconds per picture. It does not affect the application of our method in reality.

    \begin{table*}[htbp]
    \caption{The parameters, FLOPs, inference speed and results in NWPU \citep{b49} of various models.}
    \begin{center}
    \begin{tabular}{llllll}
    \hline Method & Backbone &  Parameters(M) & FLOPs(G) & Inference speed(s) & MAE In NWPU\\
    \hline 
    MCNN \citep{b18} & FS & $\textbf{0.13}$ & $\textbf{7.05}$ & $\textbf{0.008}$ & 232.5\\
    PCC-Net\citep{b52} & FS & $0.51$ & $43.87$ & $0.013$ & 167.4 \\
    CSR-Net\citep{b44} & VGG16 & $16.26$ & $108.34$ & $0.038$ & 121.3\\
    CAN\citep{b22} & VGG16 & $18.10$ & $114.83$ & $0.047$ & 106.3\\
    SCAR\citep{b45} & VGG16 & $16.29$ & $108.44$ & $0.047$ & 110.0\\
    SFANet\citep{b24} & VGG16 & $15.92$ & $93.27$ & $0.043$& $-$ \\
    M-SFANet\citep{b28} & VGG16 & $22.88$ & $115.14$ & $0.058$& $-$ \\
    SFCN\citep{b48} & ResNet101 & $38.60$ & $162.03$ & $0.096$& 105.7 \\
    RANet(ours) & VGG16 & $22.88$ & $205.97$ & $0.095$ & $\textbf{77.5}$\\
    \hline
    \end{tabular}
    \end{center} 
    \label{tab6}
    Note: The parameters and FLOPs are computes with the input size of 512×512 on a single NVIDIA 3090 GPU. The inference time is the average time of 100 runs on testing 1024×768 sample. “FS” represents that the model is trained From Scratch.
    \end{table*}

\begin{table*}[htbp]
\setlength\tabcolsep{3pt}
\caption{The parameters and inference speeds of two models. The parameters are computes with the input size of 512×512 on a single NVIDIA 3090 GPU. The inference time is the average time of 100 runs on testing 1024×768 sample.}
\begin{center}
\begin{tabular}{lllll}
\hline Method & Parameters(M) & Inference speed(s) & ShanghaiTechA  & ShanghaiTechB \\
 & & & (MAE$/$MSE)&(MAE$/$MSE)  \\
\hline VGG16+PFN & $15.86$ & $0.041$ & $68.01/109.94$ & $7.48/12.08$ \\
VGG16+PFN+RA (OURS) & $15.87$ & $0.076$ & $64.84/103.58$ & $6.30/10.14$ \\
\hline
\end{tabular}
\end{center} 
\label{tab7}
\end{table*}

To better evaluate the complexity of our method, we simply employ VGG16 with Feature Pyramid fusion like U-Net as backbone. We add our proposed feedback architecture and Region-Aware block on the VGG16 backbone. We named these two models as VGG16+PFN and VGG16+PFN+RA. The quantitative results of VGG16+FPN and VGG16+FPN +RA on ShanghaiTech have been reported in Table \ref{tab7}. As shown in Table \ref{tab7}, our VGG16+PFN+RA model is superior to the base model VGG16+PFN. More specifically, on the ShanghaiTech A, our model reduces the MAE and MSE, from 68.01 to 64.84 in MAE and from 109.94 to 103.58 in MSE. On the ShanghaiTech B, our model reduces the MAE and MSE, from 7.48 to 6.30 in MAE and from 12.08 to 10.14 in MSE. Compared to the improvement of the result, it is acceptable that our method has an increase the inference time.

\subsubsection{normal CNN networks for scale variation}
To verity the statement that normal CNN networks may have limited effects for rapid scale changes in complex scenes as they may assign a fixed scale for large objects, we simply employ two models named VGG16 and VGG16+SFN respectively to conduct experiments on ShanghaiTech dataset and UCF-QNRF dataset.
For VGG16, we simply employ the first pretrained 13 layers of VGG16 with batch normalization as encoder header. Then, we put the output of the backbone to a decoder header which consists of three 3×3 convolutional layers with 256, 64 and 32 channels respectively, and 1×1 convolutional layers to get final density map. While for VGG16+FPN, considering that different layers may focus on different abstract level features, we up-sample and cascade these multiple features as Feature Pyramid Networks.

\begin{table}[htbp]
\setlength\tabcolsep{2pt}
\caption{Quantitative results of VGG16 and VGG16+FPN in ShanghaiTech and UCF-QNRF.}
\begin{center}
\begin{tabular}{llll}
\hline \multirow{2}{*}{Method} & ShanghaiTech A & ShanghaiTech B & UCF-QNRF \\
  &(MAE$/$MSE) &(MAE$/$MSE) & (MAE$/$MSE) \\
\hline VGG16 & $67.64 / 109.46$ & $7.70 / 12.55$ & $89.52 / 154.60$ \\
VGG16+PFN & $68.01 / 109.94$ & $7.48 / 12.08$ & $88.48 / 155.84$ \\
\hline
\end{tabular}
\end{center}
\label{tab8}
\end{table}

As can be seen from Table \ref{tab8}, the normal network VGG16 is able to achieve a similar performance with VGG16+FPN which fuses multiple features of different layers in ShanghaiTech A and UCF-QNRF datasets. The reason may be that features of low layers contribute less to crowd counting in complex dense scenes. While in ShanghaiTech B which contains less people and scale variation are not prominent, VGG16+FPN with fusing multiple features of different layers performs better than normal VGG16.
These results further indicate that normal networks may have limited effects for rapid scale variation in complex scenes.

\subsubsection{Comparison of different boost strategies}

The work \citep{b53} shows that, Deep Convolutional Neural Networks (DCNNs) and human visual system are not necessarily equivalent models in object recognition. DCNNs likely achieve an object recognition competence through a set of mechanisms that are distinct from those in humans \citep{b53}. As a result, the patch-based and superpixel-based operations which are common and consistent with human visual system,  may not be optimal choices for some applications.
There are some column-based methods, which are applied in 3D point cloud \citep{b57}, localization task \citep{b58} and classification task \citep{b59} and have achieved good performances. In order to explore which method is suitable for our dense crowd counting task, we compare the column-based operation with patch-based and superpixel-based operations from theoretical analysis and experimental verification.

Firstly, column-based operation is able to boost the important information of the image. If two vectors are similar, then they will get a high score and will be boosted in our column-based method. In dense crowd counting task, the crowd areas occupy most of the content so that the crowd regions would be highlighted after column-based operation. As a result, column-based operation would boost the crowd regions and would not destroy the semantic information inside the image.
Secondly, dividing the dense image into patches for operation may divide the crowd into different areas which may cause the separation of semantic information. Moreover, patch-based operation is somewhat similar to convolution, and the extracted information may be similar to the features obtained by convolution.
Finally, superpixel-based operation focus on individual pixels which may be a lack of consideration of context in dense scenes.

We perform ablation studies on ShanghaiTech A dataset in Table \ref{tab9} to evaluate the effectiveness of each proposed component. In superpixel-based method, we do superpixel-based operation between input images and obtained maps. For patch-based method, we divide the input images and maps into $16 \times 16$ patches, then we do similarity measurement and relevance embedding for these patches as we introduced in Region-Aware Block.
Comparing base model with superpixel-based method, patch-based method and our column-based method, we find that using feedback to enhance inputs could improve the performance of base model. As can be seen from Table \ref{tab9}, our column-based method performs better than common superpixel-based method and patch-based method in dense scenes. This further evaluates the effectiveness of our proposed Region-Aware block.

\begin{figure*}[htbp]
\centering
\centerline{\includegraphics[scale=0.35]{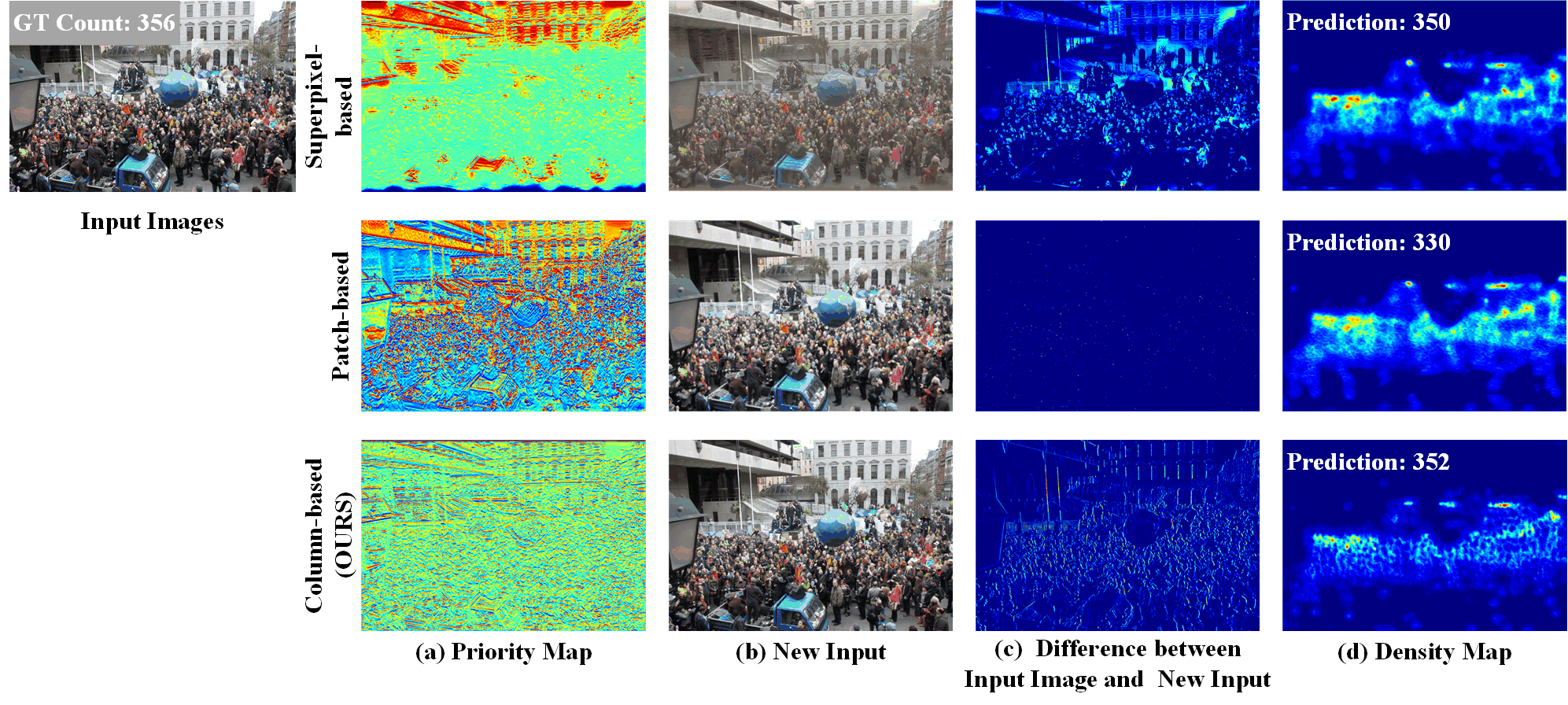}}
\caption{Visualization of outputs of superpixel-based method, patch-based method and our column-based method. (a) is the corresponding priority maps; (b) is the new inputs obtained from the networks; (c) is obtained by subtracting input image from (b) new inputs; (d) is the predicted density maps.}
\label{fig9}
\end{figure*}

\begin{table}[htbp]
\begin{center}
\caption{Comparison of different boost strategies. Superpixel-based method mean that we do superpixel-based operation between input images and obtained maps. For patch-based method, we divide the input images and maps into patches, then we do similarity measurement and relevance embedding for these patches as we introduced in Region-Aware Block.}
\begin{tabular}{lll}
\hline Method & \multicolumn{2}{c}{ ShanghaiTech $\mathrm{A}$} \\
& MAE & MSE \\
\hline Base Model & $65.84$ & $110.22$ \\
superpixel-based & $60.13$ & $100.34$ \\
patch-based (16×16) & $64.65$ & $106.79$ \\
column-based (OURS) & $57.92$ & $99.23$ \\
\hline
\end{tabular}
\label{tab9}
\end{center}
\end{table}

To further explore the difference of different boost strategies, we visualize the outputs of these three models in Fig. \ref{fig9}. We respectively visualize the superpixel-based method, patch-based method and our column-based method in the first row, second row and third row. Comparing the differences between input image and new input, we find that superpixel-based method and our column-based method could effectively boost the dense crowd regions. This verifies that our column-based method could boost the important information in input images in dense scenes. While for patch-based method, it has limited effect on boosting the crowd regions by observing the difference between input image and new input. The reason may be that the extracted feature of patch-based method may be similar to the features extracted by convolution.
Moreover, the obtained new input of column-based method is similar to input image. This further proves that our column-based method would not destroy the semantic information inside the image. As shown in the third column and fourth column in Fig. \ref{fig9},we observe that our column-based method could extract more reliable detailed edge texture feature from global receptive field and performs better than superpixel-based method in dense scenes which indicates our column-based method could consider more contextual information than superpixel-based method in dense scenes.

\subsubsection{Visualization of the density maps of base model and RANet}

\begin{figure*}[htbp]
\centering
\centerline{\includegraphics[scale=0.55]{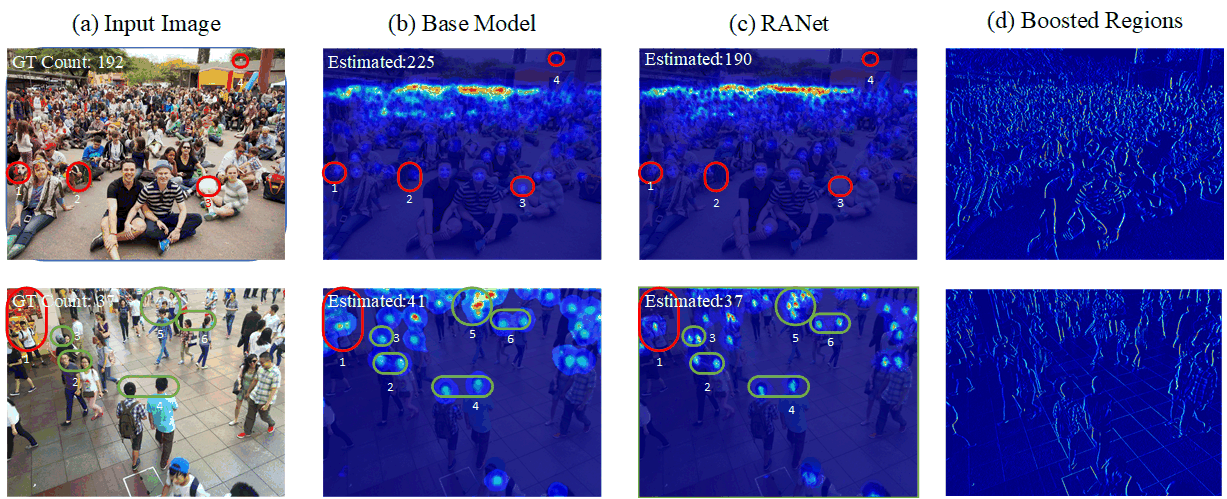}}
\caption{Visualization of the density maps of base model and RANet.}
\label{fig10}
\end{figure*}

We first introduce a novel feedback architecture to generate priority map focusing on crowd regions. Then we scan the images in the form of column vector to obtain global contextual information and boost the crowd regions in dense scenes for counting according to priority maps. The priority map makes our model pay more attention to crowd regions.
And the column vector scanning could extract reliable edge and texture information of targets from global receptive field. The combination of priority map and column vector scanning could synergistically boost the crowd regions effectively. The boosted regions are shown in Fig. \ref{fig10} (d). Therefore, our model could pay more attention to crowd regions so that it is able to relieve the problem of background noise. As shown in the red circles in the first and the second rows of Fig. \ref{fig10}, we could observe that base model is easier to mistake background for people. While our RANet could distinguish people and background. Compared with base model, the results of our RANet have more consistent responds for crowd regions. This shows that our RANet could relieve the problem of background noise.
    
Moreover, our model could extract reliable edge and texture information of targets from global receptive field with scanning the images in the form of column vector. As shown in Fig. \ref{fig10} (d), our model could extract the size of each target. As a result, our model has the ability to implicitly encode the size of each target into the feature. Thus, our region-aware block could relieve the problem of scale variation. We could also observe that the respond ranges of base model are larger than scales of targets in green circle 2, green circle 4 and green circle 6. Our RANet has different responds ranges for different targets, which reflects the scale variation. Moreover, as shown in green circle 3 and green circle 5, the density map of base model has too large responds ranges to distinguish each people. While our RANet has suitable respond ranges and is able to distinguish each target. These results further indicate that our RANet has suitable responds ranges for different targets, which reflects our method could relieve the problem of scale variation.

\section{Conclusion}

In this paper, we proposed a novel feedback architecture model with Region-Aware block modeling human's Top-Down visual perception mechanism, name RANet, aiming to deal with background noise and scale variation. Firstly, we introduce a feedback architecture to train priority map that provide prior about candidate crowd region in input images. This prior would be fully utilized in Region-Aware block to reduce background noise and capture global context information. More specifically, we scan the whole input images and its priority maps in the form of column vector to obtain a relevance matrix for measuring their similarity. The relevance matrix obtained would be utilized to build global relationships between pixels. In other word, the Region-Aware block could adaptively encode the contextual information into input images through global receptive field. So that the RANet shows powerful ability to estimate different dense scenes through attention the crowd regions and the congestion degree of crowd regions with a global receptive filed.







\section*{Acknowledgment}
This work was supported by the National Natural Science Foundation of China under Grant Nos. 62073257, 62141223, U21A20485 and 61971343, and the Natural Science Basic Research Plan in Shaanxi Province of China under Grant No. 2020JM-012.
\printcredits

\bibliographystyle{model5-names}

\bibliography{manuscript}

\begin{thebibliography}{48}
\expandafter\ifx\csname natexlab\endcsname\relax\def\natexlab#1{#1}\fi
\providecommand{\url}[1]{\texttt{#1}}
\providecommand{\href}[2]{#2}
\providecommand{\path}[1]{#1}
\providecommand{\DOIprefix}{doi:}
\providecommand{\ArXivprefix}{arXiv:}
\providecommand{\URLprefix}{URL: }
\providecommand{\Pubmedprefix}{pmid:}
\providecommand{\doi}[1]{\href{http://dx.doi.org/#1}{\path{#1}}}
\providecommand{\Pubmed}[1]{\href{pmid:#1}{\path{#1}}}
\providecommand{\bibinfo}[2]{#2}
\ifx\xfnm\relax \def\xfnm[#1]{\unskip,\space#1}\fi
\bibitem[{Ali et~al.(2022)Ali, Zhu \& Zakarya}]{b60}
\bibinfo{author}{Ali, A.}, \bibinfo{author}{Zhu, Y.}, \&
  \bibinfo{author}{Zakarya, M.} (\bibinfo{year}{2022}).
\newblock \bibinfo{title}{Exploiting dynamic spatio-temporal graph
  convolutional neural networks for citywide traffic flows prediction}.
\newblock {\it \bibinfo{journal}{Neural networks}\/},  {\it
  \bibinfo{volume}{145}\/}, \bibinfo{pages}{233--247}.
\bibitem[{Babu~Sam et~al.(2017)Babu~Sam, Surya \& Venkatesh~Babu}]{b20}
\bibinfo{author}{Babu~Sam, D.}, \bibinfo{author}{Surya, S.}, \&
  \bibinfo{author}{Venkatesh~Babu, R.} (\bibinfo{year}{2017}).
\newblock \bibinfo{title}{Switching convolutional neural network for crowd
  counting}.
\newblock In {\it \bibinfo{booktitle}{Proceedings of the IEEE conference on
  computer vision and pattern recognition}\/} (pp.
  \bibinfo{pages}{5744--5752}).
\bibitem[{Bansal \& Venkatesh(2015)}]{b36}
\bibinfo{author}{Bansal, A.}, \& \bibinfo{author}{Venkatesh, K.~S.}
  (\bibinfo{year}{2015}).
\newblock \bibinfo{title}{People counting in high density crowds from still
  images}.
\newblock \href{http://arxiv.org/abs/1507.08445}{\tt arXiv:1507.08445}.
\bibitem[{Boominathan et~al.(2016)Boominathan, Kruthiventi \& Babu}]{b17}
\bibinfo{author}{Boominathan, L.}, \bibinfo{author}{Kruthiventi, S.~S.}, \&
  \bibinfo{author}{Babu, R.~V.} (\bibinfo{year}{2016}).
\newblock \bibinfo{title}{Crowdnet: A deep convolutional network for dense
  crowd counting}.
\newblock In {\it \bibinfo{booktitle}{Proceedings of the 24th ACM international
  conference on Multimedia}\/} (pp. \bibinfo{pages}{640--644}).
\bibitem[{Cao et~al.(2018)Cao, Wang, Zhao \& Su}]{b45}
\bibinfo{author}{Cao, X.}, \bibinfo{author}{Wang, Z.}, \bibinfo{author}{Zhao,
  Y.}, \& \bibinfo{author}{Su, F.} (\bibinfo{year}{2018}).
\newblock \bibinfo{title}{Scale aggregation network for accurate and efficient
  crowd counting}.
\newblock In {\it \bibinfo{booktitle}{Proceedings of the European Conference on
  Computer Vision (ECCV)}\/} (pp. \bibinfo{pages}{734--750}).
\bibitem[{Chan \& Vasconcelos(2009)}]{b14}
\bibinfo{author}{Chan, A.~B.}, \& \bibinfo{author}{Vasconcelos, N.}
  (\bibinfo{year}{2009}).
\newblock \bibinfo{title}{Bayesian poisson regression for crowd counting}.
\newblock In {\it \bibinfo{booktitle}{IEEE 12th international conference on
  computer vision}\/} (pp. \bibinfo{pages}{545--551}).
\newblock \bibinfo{organization}{IEEE}.
\bibitem[{Chen et~al.(2017)Chen, Papandreou, Kokkinos, Murphy \& Yuille}]{b33}
\bibinfo{author}{Chen, L.-C.}, \bibinfo{author}{Papandreou, G.},
  \bibinfo{author}{Kokkinos, I.}, \bibinfo{author}{Murphy, K.}, \&
  \bibinfo{author}{Yuille, A.~L.} (\bibinfo{year}{2017}).
\newblock \bibinfo{title}{Deeplab: Semantic image segmentation with deep
  convolutional nets, atrous convolution, and fully connected crfs}.
\newblock {\it \bibinfo{journal}{IEEE transactions on pattern analysis and
  machine intelligence}\/},  {\it \bibinfo{volume}{40}\/},
  \bibinfo{pages}{834--848}.
\bibitem[{Dalal \& Triggs()}]{b10}
\bibinfo{author}{Dalal, N.}, \& \bibinfo{author}{Triggs, B.} ().
\newblock \bibinfo{title}{Histograms of oriented gradients for human
  detection}.
\newblock In {\it \bibinfo{booktitle}{2005 IEEE computer society conference on
  computer vision and pattern recognition (CVPR'05)}\/} (pp.
  \bibinfo{pages}{886--893}).
\newblock \bibinfo{organization}{Ieee} volume~\bibinfo{volume}{1}.
\bibitem[{Dollar et~al.(2011)Dollar, Wojek, Schiele \& Perona}]{b9}
\bibinfo{author}{Dollar, P.}, \bibinfo{author}{Wojek, C.},
  \bibinfo{author}{Schiele, B.}, \& \bibinfo{author}{Perona, P.}
  (\bibinfo{year}{2011}).
\newblock \bibinfo{title}{Pedestrian detection: An evaluation of the state of
  the art}.
\newblock {\it \bibinfo{journal}{IEEE transactions on pattern analysis and
  machine intelligence}\/},  {\it \bibinfo{volume}{34}\/},
  \bibinfo{pages}{743--761}.
\bibitem[{Fiaschi et~al.(2012)Fiaschi, K{\"o}the, Nair \& Hamprecht}]{b15}
\bibinfo{author}{Fiaschi, L.}, \bibinfo{author}{K{\"o}the, U.},
  \bibinfo{author}{Nair, R.}, \& \bibinfo{author}{Hamprecht, F.~A.}
  (\bibinfo{year}{2012}).
\newblock \bibinfo{title}{Learning to count with regression forest and
  structured labels}.
\newblock In {\it \bibinfo{booktitle}{Proceedings of the 21st International
  Conference on Pattern Recognition}\/} (pp. \bibinfo{pages}{2685--2688}).
\newblock \bibinfo{organization}{IEEE}.
\bibitem[{Gao et~al.(2019{\natexlab{a}})Gao, Wang \& Li}]{b52}
\bibinfo{author}{Gao, J.}, \bibinfo{author}{Wang, Q.}, \& \bibinfo{author}{Li,
  X.} (\bibinfo{year}{2019}{\natexlab{a}}).
\newblock \bibinfo{title}{Pcc net: Perspective crowd counting via spatial
  convolutional network}.
\newblock {\it \bibinfo{journal}{IEEE Transactions on Circuits and Systems for
  Video Technology}\/},  {\it \bibinfo{volume}{30}\/},
  \bibinfo{pages}{3486--3498}.
\bibitem[{Gao et~al.(2019{\natexlab{b}})Gao, Wang \& Yuan}]{b39}
\bibinfo{author}{Gao, J.}, \bibinfo{author}{Wang, Q.}, \&
  \bibinfo{author}{Yuan, Y.} (\bibinfo{year}{2019}{\natexlab{b}}).
\newblock \bibinfo{title}{Scar: Spatial-/channel-wise attention regression
  networks for crowd counting}.
\newblock {\it \bibinfo{journal}{Neurocomputing}\/},  {\it
  \bibinfo{volume}{363}\/}, \bibinfo{pages}{1--8}.
\bibitem[{Hossain et~al.()Hossain, Hosseinzadeh, Chanda \& Wang}]{b27}
\bibinfo{author}{Hossain, M.}, \bibinfo{author}{Hosseinzadeh, M.},
  \bibinfo{author}{Chanda, O.}, \& \bibinfo{author}{Wang, Y.} ().
\newblock \bibinfo{title}{Crowd counting using scale-aware attention networks}.
\newblock In {\it \bibinfo{booktitle}{2019 IEEE Winter Conference on
  Applications of Computer Vision (WACV)}\/} (pp. \bibinfo{pages}{1280--1288}).
\newblock \bibinfo{organization}{IEEE}.
\bibitem[{Idrees et~al.(2018)Idrees, Tayyab, Athrey, Zhang, Al-Maadeed, Rajpoot
  \& Shah}]{b35}
\bibinfo{author}{Idrees, H.}, \bibinfo{author}{Tayyab, M.},
  \bibinfo{author}{Athrey, K.}, \bibinfo{author}{Zhang, D.},
  \bibinfo{author}{Al-Maadeed, S.}, \bibinfo{author}{Rajpoot, N.}, \&
  \bibinfo{author}{Shah, M.} (\bibinfo{year}{2018}).
\newblock \bibinfo{title}{Composition loss for counting, density map estimation
  and localization in dense crowds}.
\newblock In {\it \bibinfo{booktitle}{Proceedings of the European Conference on
  Computer Vision (ECCV)}\/} (pp. \bibinfo{pages}{532--546}).
\bibitem[{Jiang et~al.(2020)Jiang, Zhang, Xu, Zhang, Lv, Zhou, Yang \&
  Pang}]{b5}
\bibinfo{author}{Jiang, X.}, \bibinfo{author}{Zhang, L.}, \bibinfo{author}{Xu,
  M.}, \bibinfo{author}{Zhang, T.}, \bibinfo{author}{Lv, P.},
  \bibinfo{author}{Zhou, B.}, \bibinfo{author}{Yang, X.}, \&
  \bibinfo{author}{Pang, Y.} (\bibinfo{year}{2020}).
\newblock \bibinfo{title}{Attention scaling for crowd counting}.
\newblock In {\it \bibinfo{booktitle}{Proceedings of the IEEE/CVF Conference on
  Computer Vision and Pattern Recognition}\/} (pp.
  \bibinfo{pages}{4706--4715}).
\bibitem[{Kim \& Kim(2018)}]{b57}
\bibinfo{author}{Kim, G.}, \& \bibinfo{author}{Kim, A.} (\bibinfo{year}{2018}).
\newblock \bibinfo{title}{Scan context: Egocentric spatial descriptor for place
  recognition within 3d point cloud map}.
\newblock In {\it \bibinfo{booktitle}{IEEE/RSJ International Conference on
  Intelligent Robots and Systems (IROS)}\/} (pp. \bibinfo{pages}{4802--4809}).
\newblock \bibinfo{organization}{IEEE}.
\bibitem[{Kim et~al.(2019)Kim, Park \& Kim}]{b58}
\bibinfo{author}{Kim, G.}, \bibinfo{author}{Park, B.}, \& \bibinfo{author}{Kim,
  A.} (\bibinfo{year}{2019}).
\newblock \bibinfo{title}{1-day learning, 1-year localization: Long-term lidar
  localization using scan context image}.
\newblock {\it \bibinfo{journal}{IEEE Robotics and Automation Letters}\/},
  {\it \bibinfo{volume}{4}\/}, \bibinfo{pages}{1948--1955}.
\bibitem[{Li et~al.(2018)Li, Zhang \& Chen}]{b44}
\bibinfo{author}{Li, Y.}, \bibinfo{author}{Zhang, X.}, \&
  \bibinfo{author}{Chen, D.} (\bibinfo{year}{2018}).
\newblock \bibinfo{title}{Csrnet: Dilated convolutional neural networks for
  understanding the highly congested scenes}.
\newblock In {\it \bibinfo{booktitle}{Proceedings of the IEEE conference on
  computer vision and pattern recognition}\/} (pp.
  \bibinfo{pages}{1091--1100}).
\bibitem[{Liu et~al.(2018)Liu, Gao, Meng \& Hauptmann}]{b26}
\bibinfo{author}{Liu, J.}, \bibinfo{author}{Gao, C.}, \bibinfo{author}{Meng,
  D.}, \& \bibinfo{author}{Hauptmann, A.~G.} (\bibinfo{year}{2018}).
\newblock \bibinfo{title}{Decidenet: Counting varying density crowds through
  attention guided detection and density estimation}.
\newblock In {\it \bibinfo{booktitle}{Proceedings of the IEEE Conference on
  Computer Vision and Pattern Recognition}\/} (pp.
  \bibinfo{pages}{5197--5206}).
\bibitem[{Liu et~al.(2019{\natexlab{a}})Liu, Long, Zou, Niu, Pan \& Wu}]{b2}
\bibinfo{author}{Liu, N.}, \bibinfo{author}{Long, Y.}, \bibinfo{author}{Zou,
  C.}, \bibinfo{author}{Niu, Q.}, \bibinfo{author}{Pan, L.}, \&
  \bibinfo{author}{Wu, H.} (\bibinfo{year}{2019}{\natexlab{a}}).
\newblock \bibinfo{title}{Adcrowdnet: An attention-injective deformable
  convolutional network for crowd understanding}.
\newblock In {\it \bibinfo{booktitle}{Proceedings of the IEEE/CVF Conference on
  Computer Vision and Pattern Recognition}\/} (pp.
  \bibinfo{pages}{3225--3234}).
\bibitem[{Liu et~al.(2019{\natexlab{b}})Liu, Salzmann \& Fua}]{b22}
\bibinfo{author}{Liu, W.}, \bibinfo{author}{Salzmann, M.}, \&
  \bibinfo{author}{Fua, P.} (\bibinfo{year}{2019}{\natexlab{b}}).
\newblock \bibinfo{title}{Context-aware crowd counting}.
\newblock In {\it \bibinfo{booktitle}{Proceedings of the IEEE/CVF Conference on
  Computer Vision and Pattern Recognition}\/} (pp.
  \bibinfo{pages}{5099--5108}).
\bibitem[{Lonnqvist et~al.(2020)Lonnqvist, Clarke \& Chakravarthi}]{b53}
\bibinfo{author}{Lonnqvist, B.}, \bibinfo{author}{Clarke, A.~D.}, \&
  \bibinfo{author}{Chakravarthi, R.} (\bibinfo{year}{2020}).
\newblock \bibinfo{title}{Crowding in humans is unlike that in convolutional
  neural networks}.
\newblock {\it \bibinfo{journal}{Neural Networks}\/},  {\it
  \bibinfo{volume}{126}\/}, \bibinfo{pages}{262--274}.
\bibitem[{Ma et~al.(2019)Ma, Wei, Hong \& Gong}]{b34}
\bibinfo{author}{Ma, Z.}, \bibinfo{author}{Wei, X.}, \bibinfo{author}{Hong,
  X.}, \& \bibinfo{author}{Gong, Y.} (\bibinfo{year}{2019}).
\newblock \bibinfo{title}{Bayesian loss for crowd count estimation with point
  supervision}.
\newblock In {\it \bibinfo{booktitle}{Proceedings of the IEEE/CVF International
  Conference on Computer Vision}\/} (pp. \bibinfo{pages}{6142--6151}).
\bibitem[{Onoro-Rubio \& L{\'o}pez-Sastre(2016{\natexlab{a}})}]{b1}
\bibinfo{author}{Onoro-Rubio, D.}, \& \bibinfo{author}{L{\'o}pez-Sastre, R.~J.}
  (\bibinfo{year}{2016}{\natexlab{a}}).
\newblock \bibinfo{title}{Towards perspective-free object counting with deep
  learning}.
\newblock In {\it \bibinfo{booktitle}{European conference on computer
  vision}\/} (pp. \bibinfo{pages}{615--629}).
\newblock \bibinfo{organization}{Springer}.
\bibitem[{Onoro-Rubio \& L{\'o}pez-Sastre(2016{\natexlab{b}})}]{b23}
\bibinfo{author}{Onoro-Rubio, D.}, \& \bibinfo{author}{L{\'o}pez-Sastre, R.~J.}
  (\bibinfo{year}{2016}{\natexlab{b}}).
\newblock \bibinfo{title}{Towards perspective-free object counting with deep
  learning}.
\newblock In {\it \bibinfo{booktitle}{European conference on computer
  vision}\/} (pp. \bibinfo{pages}{615--629}).
\newblock \bibinfo{organization}{Springer}.
\bibitem[{Parmar et~al.(2018)Parmar, Vaswani, Uszkoreit, Kaiser, Shazeer, Ku \&
  Tran}]{b56}
\bibinfo{author}{Parmar, N.}, \bibinfo{author}{Vaswani, A.},
  \bibinfo{author}{Uszkoreit, J.}, \bibinfo{author}{Kaiser, L.},
  \bibinfo{author}{Shazeer, N.}, \bibinfo{author}{Ku, A.}, \&
  \bibinfo{author}{Tran, D.} (\bibinfo{year}{2018}).
\newblock \bibinfo{title}{Image transformer}.
\newblock In {\it \bibinfo{booktitle}{International Conference on Machine
  Learning}\/} (pp. \bibinfo{pages}{4055--4064}).
\newblock \bibinfo{organization}{PMLR}.
\bibitem[{Rodriguez-Vazquez et~al.(2022)Rodriguez-Vazquez, Alvarez-Fernandez,
  Molina \& Campoy}]{b61}
\bibinfo{author}{Rodriguez-Vazquez, J.}, \bibinfo{author}{Alvarez-Fernandez,
  A.}, \bibinfo{author}{Molina, M.}, \& \bibinfo{author}{Campoy, P.}
  (\bibinfo{year}{2022}).
\newblock \bibinfo{title}{Zenithal isotropic object counting by localization
  using adversarial training}.
\newblock {\it \bibinfo{journal}{Neural Networks}\/},  {\it
  \bibinfo{volume}{145}\/}, \bibinfo{pages}{155--163}.
\bibitem[{Rong \& Li(2021)}]{b3}
\bibinfo{author}{Rong, L.}, \& \bibinfo{author}{Li, C.} (\bibinfo{year}{2021}).
\newblock \bibinfo{title}{Coarse-and fine-grained attention network with
  background-aware loss for crowd density map estimation}.
\newblock In {\it \bibinfo{booktitle}{Proceedings of the IEEE/CVF Winter
  Conference on Applications of Computer Vision}\/} (pp.
  \bibinfo{pages}{3675--3684}).
\bibitem[{Ryan et~al.(2009)Ryan, Denman, Fookes \& Sridharan}]{b13}
\bibinfo{author}{Ryan, D.}, \bibinfo{author}{Denman, S.},
  \bibinfo{author}{Fookes, C.}, \& \bibinfo{author}{Sridharan, S.}
  (\bibinfo{year}{2009}).
\newblock \bibinfo{title}{Crowd counting using multiple local features}.
\newblock In {\it \bibinfo{booktitle}{Digital Image Computing: Techniques and
  Applications}\/} (pp. \bibinfo{pages}{81--88}).
\newblock \bibinfo{organization}{IEEE}.
\bibitem[{Sam et~al.(2020)Sam, Peri, Sundararaman, Kamath \&
  Radhakrishnan}]{b47}
\bibinfo{author}{Sam, D.~B.}, \bibinfo{author}{Peri, S.~V.},
  \bibinfo{author}{Sundararaman, M.~N.}, \bibinfo{author}{Kamath, A.}, \&
  \bibinfo{author}{Radhakrishnan, V.~B.} (\bibinfo{year}{2020}).
\newblock \bibinfo{title}{Locate, size and count: Accurately resolving people
  in dense crowds via detection}.
\newblock {\it \bibinfo{journal}{IEEE transactions on pattern analysis and
  machine intelligence}\/},  {\it \bibinfo{volume}{PP}\/},
  \bibinfo{pages}{1--1}.
\bibitem[{Shang et~al.(2016)Shang, Ai \& Bai}]{b19}
\bibinfo{author}{Shang, C.}, \bibinfo{author}{Ai, H.}, \& \bibinfo{author}{Bai,
  B.} (\bibinfo{year}{2016}).
\newblock \bibinfo{title}{End-to-end crowd counting via joint learning local
  and global count}.
\newblock In {\it \bibinfo{booktitle}{IEEE International Conference on Image
  Processing (ICIP)}\/} (pp. \bibinfo{pages}{1215--1219}).
\newblock \bibinfo{organization}{IEEE}.
\bibitem[{Sindagi et~al.(2020)Sindagi, Yasarla \& Patel}]{b37}
\bibinfo{author}{Sindagi, V.}, \bibinfo{author}{Yasarla, R.}, \&
  \bibinfo{author}{Patel, V.~M.} (\bibinfo{year}{2020}).
\newblock \bibinfo{title}{Jhu-crowd++: Large-scale crowd counting dataset and a
  benchmark method}.
\newblock {\it \bibinfo{journal}{IEEE Transactions on Pattern Analysis and
  Machine Intelligence}\/},  (pp. \bibinfo{pages}{1--1}).
\bibitem[{Sindagi \& Patel(2017)}]{b21}
\bibinfo{author}{Sindagi, V.~A.}, \& \bibinfo{author}{Patel, V.~M.}
  (\bibinfo{year}{2017}).
\newblock \bibinfo{title}{Cnn-based cascaded multi-task learning of high-level
  prior and density estimation for crowd counting}.
\newblock In {\it \bibinfo{booktitle}{14th IEEE International Conference on
  Advanced Video and Signal Based Surveillance (AVSS)}\/} (pp.
  \bibinfo{pages}{1--6}).
\newblock \bibinfo{organization}{IEEE}.
\bibitem[{Sindagi \& Patel(2019)}]{b46}
\bibinfo{author}{Sindagi, V.~A.}, \& \bibinfo{author}{Patel, V.~M.}
  (\bibinfo{year}{2019}).
\newblock \bibinfo{title}{Multi-level bottom-top and top-bottom feature fusion
  for crowd counting}.
\newblock In {\it \bibinfo{booktitle}{Proceedings of the IEEE/CVF International
  Conference on Computer Vision}\/} (pp. \bibinfo{pages}{1002--1012}).
\bibitem[{Szeskin et~al.(2021)Szeskin, Yehuda, Shmueli, Levy \&
  Joskowicz}]{b59}
\bibinfo{author}{Szeskin, A.}, \bibinfo{author}{Yehuda, R.},
  \bibinfo{author}{Shmueli, O.}, \bibinfo{author}{Levy, J.}, \&
  \bibinfo{author}{Joskowicz, L.} (\bibinfo{year}{2021}).
\newblock \bibinfo{title}{A column-based deep learning method for the detection
  and quantification of atrophy associated with amd in oct scans}.
\newblock {\it \bibinfo{journal}{Medical Image Analysis}\/},  (p.
  \bibinfo{pages}{102130}).
\bibitem[{Thanasutives et~al.(2021)Thanasutives, Fukui, Numao \&
  Kijsirikul}]{b28}
\bibinfo{author}{Thanasutives, P.}, \bibinfo{author}{Fukui, K.-i.},
  \bibinfo{author}{Numao, M.}, \& \bibinfo{author}{Kijsirikul, B.}
  (\bibinfo{year}{2021}).
\newblock \bibinfo{title}{Encoder-decoder based convolutional neural networks
  with multi-scale-aware modules for crowd counting}.
\newblock In {\it \bibinfo{booktitle}{25th International Conference on Pattern
  Recognition (ICPR)}\/} (pp. \bibinfo{pages}{2382--2389}).
\newblock \bibinfo{organization}{IEEE}.
\bibitem[{Viola \& Jones(2004)}]{b11}
\bibinfo{author}{Viola, P.}, \& \bibinfo{author}{Jones, M.~J.}
  (\bibinfo{year}{2004}).
\newblock \bibinfo{title}{Robust real-time face detection}.
\newblock {\it \bibinfo{journal}{International journal of computer vision}\/},
  {\it \bibinfo{volume}{57}\/}, \bibinfo{pages}{137--154}.
\bibitem[{Wang et~al.(2020{\natexlab{a}})Wang, Liu, Samaras \& Hoai}]{b50}
\bibinfo{author}{Wang, B.}, \bibinfo{author}{Liu, H.},
  \bibinfo{author}{Samaras, D.}, \& \bibinfo{author}{Hoai, M.}
  (\bibinfo{year}{2020}{\natexlab{a}}).
\newblock \bibinfo{title}{Distribution matching for crowd counting}.
\newblock In {\it \bibinfo{booktitle}{Advances in Neural Information Processing
  Systems}\/}.
\bibitem[{Wang et~al.(2020{\natexlab{b}})Wang, Gao, Lin \& Li}]{b49}
\bibinfo{author}{Wang, Q.}, \bibinfo{author}{Gao, J.}, \bibinfo{author}{Lin,
  W.}, \& \bibinfo{author}{Li, X.} (\bibinfo{year}{2020}{\natexlab{b}}).
\newblock \bibinfo{title}{Nwpu-crowd: A large-scale benchmark for crowd
  counting and localization}.
\newblock {\it \bibinfo{journal}{IEEE transactions on pattern analysis and
  machine intelligence}\/},  {\it \bibinfo{volume}{43}\/},
  \bibinfo{pages}{2141--2149}.
\bibitem[{Wang et~al.(2019)Wang, Gao, Lin \& Yuan}]{b48}
\bibinfo{author}{Wang, Q.}, \bibinfo{author}{Gao, J.}, \bibinfo{author}{Lin,
  W.}, \& \bibinfo{author}{Yuan, Y.} (\bibinfo{year}{2019}).
\newblock \bibinfo{title}{Learning from synthetic data for crowd counting in
  the wild}.
\newblock In {\it \bibinfo{booktitle}{Proceedings of the IEEE/CVF Conference on
  Computer Vision and Pattern Recognition}\/} (pp.
  \bibinfo{pages}{8198--8207}).
\bibitem[{Wang et~al.(2020{\natexlab{c}})Wang, Lv, Zhao, Yang \& Ruan}]{b8}
\bibinfo{author}{Wang, X.}, \bibinfo{author}{Lv, R.}, \bibinfo{author}{Zhao,
  Y.}, \bibinfo{author}{Yang, T.}, \& \bibinfo{author}{Ruan, Q.}
  (\bibinfo{year}{2020}{\natexlab{c}}).
\newblock \bibinfo{title}{Multi-scale context aggregation network with
  attention-guided for crowd counting}.
\newblock In {\it \bibinfo{booktitle}{15th IEEE International Conference on
  Signal Processing (ICSP)}\/} (pp. \bibinfo{pages}{240--245}).
\newblock \bibinfo{organization}{IEEE} volume~\bibinfo{volume}{1}.
\bibitem[{Wu \& Nevatia(2005)}]{b12}
\bibinfo{author}{Wu, B.}, \& \bibinfo{author}{Nevatia, R.}
  (\bibinfo{year}{2005}).
\newblock \bibinfo{title}{Detection of multiple, partially occluded humans in a
  single image by bayesian combination of edgelet part detectors}.
\newblock In {\it \bibinfo{booktitle}{Tenth IEEE International Conference on
  Computer Vision (ICCV'05) Volume 1}\/} (pp. \bibinfo{pages}{90--97}).
\newblock \bibinfo{organization}{IEEE} volume~\bibinfo{volume}{1}.
\bibitem[{Yang et~al.(2020)Yang, Li, Wu, Su, Huang \& Sebe}]{b6}
\bibinfo{author}{Yang, Y.}, \bibinfo{author}{Li, G.}, \bibinfo{author}{Wu, Z.},
  \bibinfo{author}{Su, L.}, \bibinfo{author}{Huang, Q.}, \&
  \bibinfo{author}{Sebe, N.} (\bibinfo{year}{2020}).
\newblock \bibinfo{title}{Reverse perspective network for perspective-aware
  object counting}.
\newblock In {\it \bibinfo{booktitle}{Proceedings of the IEEE/CVF Conference on
  Computer Vision and Pattern Recognition}\/} (pp.
  \bibinfo{pages}{4374--4383}).
\bibitem[{Zhang et~al.(2019{\natexlab{a}})Zhang, Shen, Xiao, Zhu, Zhen, Cao \&
  Shao}]{b42}
\bibinfo{author}{Zhang, A.}, \bibinfo{author}{Shen, J.}, \bibinfo{author}{Xiao,
  Z.}, \bibinfo{author}{Zhu, F.}, \bibinfo{author}{Zhen, X.},
  \bibinfo{author}{Cao, X.}, \& \bibinfo{author}{Shao, L.}
  (\bibinfo{year}{2019}{\natexlab{a}}).
\newblock \bibinfo{title}{Relational attention network for crowd counting}.
\newblock In {\it \bibinfo{booktitle}{Proceedings of the IEEE/CVF International
  Conference on Computer Vision}\/} (pp. \bibinfo{pages}{6788--6797}).
\bibitem[{Zhang et~al.(2019{\natexlab{b}})Zhang, Yue, Shen, Zhu, Zhen, Cao \&
  Shao}]{b41}
\bibinfo{author}{Zhang, A.}, \bibinfo{author}{Yue, L.}, \bibinfo{author}{Shen,
  J.}, \bibinfo{author}{Zhu, F.}, \bibinfo{author}{Zhen, X.},
  \bibinfo{author}{Cao, X.}, \& \bibinfo{author}{Shao, L.}
  (\bibinfo{year}{2019}{\natexlab{b}}).
\newblock \bibinfo{title}{Attentional neural fields for crowd counting}.
\newblock In {\it \bibinfo{booktitle}{Proceedings of the IEEE/CVF International
  Conference on Computer Vision}\/} (pp. \bibinfo{pages}{5714--5723}).
\bibitem[{Zhang et~al.(2015)Zhang, Li, Wang \& Yang}]{b16}
\bibinfo{author}{Zhang, C.}, \bibinfo{author}{Li, H.}, \bibinfo{author}{Wang,
  X.}, \& \bibinfo{author}{Yang, X.} (\bibinfo{year}{2015}).
\newblock \bibinfo{title}{Cross-scene crowd counting via deep convolutional
  neural networks}.
\newblock In {\it \bibinfo{booktitle}{Proceedings of the IEEE conference on
  computer vision and pattern recognition}\/} (pp. \bibinfo{pages}{833--841}).
\bibitem[{Zhang et~al.(2016)Zhang, Zhou, Chen, Gao \& Ma}]{b18}
\bibinfo{author}{Zhang, Y.}, \bibinfo{author}{Zhou, D.}, \bibinfo{author}{Chen,
  S.}, \bibinfo{author}{Gao, S.}, \& \bibinfo{author}{Ma, Y.}
  (\bibinfo{year}{2016}).
\newblock \bibinfo{title}{Single-image crowd counting via multi-column
  convolutional neural network}.
\newblock In {\it \bibinfo{booktitle}{Proceedings of the IEEE conference on
  computer vision and pattern recognition}\/} (pp. \bibinfo{pages}{589--597}).
\bibitem[{Zhu et~al.(2019)Zhu, Zhao, Lu, Lin, Peng \& Yao}]{b24}
\bibinfo{author}{Zhu, L.}, \bibinfo{author}{Zhao, Z.}, \bibinfo{author}{Lu,
  C.}, \bibinfo{author}{Lin, Y.}, \bibinfo{author}{Peng, Y.}, \&
  \bibinfo{author}{Yao, T.} (\bibinfo{year}{2019}).
\newblock \bibinfo{title}{Dual path multi-scale fusion networks with attention
  for crowd counting}.
\newblock \href{http://arxiv.org/abs/1902.01115}{\tt arXiv:1902.01115}.

\end{thebibliography}

\bio{}
\endbio


\end{document}